\documentclass[parskip=full,11pt]{article}

\usepackage{amssymb}
\usepackage{amsthm}
\usepackage{amsmath}
\usepackage{amsmath}
\usepackage{float,booktabs}
\usepackage{threeparttable}
\usepackage{float} 
\usepackage{graphicx}
\usepackage{hyperref}
\usepackage{natbib}
\usepackage{url}
\usepackage{xcolor}
\usepackage{dsfont}
\usepackage{soul,color,colortbl}
\usepackage{array,multirow}
\usepackage{cancel}
\DeclareMathOperator*{\argmin}{arg\,min} 
\usepackage[top=1in, bottom=1.3in, left=0.875in, right=0.875in]{geometry}
\usepackage[linesnumbered,ruled]{algorithm2e}
\usepackage[labelformat=simple]{subcaption}
\usepackage{bm}
\usepackage[linesnumbered,ruled]{algorithm2e}
\usepackage{caption}
\usepackage{subcaption}

\def\spacingset#1{\renewcommand{\baselinestretch}
{#1}\small\normalsize} \spacingset{1}

\newcolumntype{P}[1]{>{\centering\arraybackslash}p{#1}}
\newcommand*{\myfont}{\fontfamily{lmss}\selectfont}
\DeclareTextFontCommand{\textpython}{\myfont}

\DeclareCaptionLabelFormat{boldnumber}{\bothIfFirst{#1}{~}\textbf{#2}}
\title{Zero-Inflated Tweedie Boosted Trees with \texttt{CatBoost} for Insurance Loss Analytics}

\author{Banghee So \thanks{Corresponding author: Department of Mathematics, Towson University, 7800 York Rd, Towson, 21204, MD, USA. Email: \texttt{bso@towson.edu}.} 
\and Emiliano A. Valdez\thanks{Department of Mathematics, University of Connecticut, 341 Mansfield Road, Storrs, CT, 06269-1009, USA. Email: \texttt{emiliano.valdez@uconn.edu}.}}

\begin{document}

\maketitle

\begin{abstract}
In this paper, we explore advanced modifications to the Tweedie regression model in order to address its limitations in modeling aggregate claims for various types of insurance such as automobile, health, and liability. Traditional Tweedie models, while effective in capturing the probability and magnitude of claims, usually fall short in accurately representing the large incidence of zero claims. Our recommended approach involves a refined modeling of the zero-claim process, together with the integration of boosting methods in order to help leverage an iterative process to enhance predictive accuracy. Despite the inherent slowdown in learning algorithms due to this iteration, several efficient implementation techniques that also help precise tuning of parameters like \texttt{XGBoost}, \texttt{LightGBM}, and \texttt{CatBoost} have emerged. Nonetheless, we chose to utilize \texttt{CatBoost}, an efficient boosting approach that effectively handles categorical and other special types of data. The core contribution of our paper is the assembly of separate modeling for zero claims and the application of tree-based boosting ensemble methods within a \texttt{CatBoost} framework, assuming that the inflated probability of zero is a function of the mean parameter. The efficacy of our enhanced Tweedie model is demonstrated through the application of an insurance telematics dataset, which presents the additional complexity of compositional feature variables. Our modeling results reveal a marked improvement in model performance, showcasing its potential to deliver more accurate predictions suitable for insurance claim analytics.
\vspace{0.8cm}

\noindent \textbf{Keywords}: Tweedie, zero-inflated models, tree-based boosting, gradient boosting, \texttt{CatBoost}, compositional features, aggregate loss models.

\end{abstract}

\newpage

\section{Introduction} \label{sec:intro}
Insurance loss analytics, which is the backbone of modern insurance risk management strategies, involves a systematic collection, processing, and analysis of observed insurance loss data to extract insights and produce predictive models to help quantify and manage risk exposure. Actuaries and risk managers have historically employed the use of data analytics to uncover patterns, identify emerging risks, and develop predictive models for assessing claim probabilities, estimating aggregate claim amounts, and deriving necessary optimal level of reserves to set aside to cover these future estimated claims. See \citet{klugman2019loss}.

The two-part frequency-severity models have historically been the norm used in insurance loss data analytics for modeling aggregate claims for pure premium. These models recognize that the frequency of claims (how often claims occur) and the severity of claims (the amount paid per claim) are often influenced by different factors and follow different distributions. The claim frequency focuses on predicting the number of claims that occur within a specified period, such as a year or a month, and often follows a count distribution model, such as Poisson or negative binomial distribution. It considers factors that influence claim occurrence, such as policyholder characteristics (e.g., age, location), policy features (e.g., coverage type, deductible), external factors (e.g., weather conditions, economic trends), and any relevant historical data. On the other hand, the severity focuses on predicting the monetary value of individual claims, i.e., how much each claim will cost the insurer. It generally follows a continuous distribution, such as gamma or log-normal, that captures the variability in claim amounts and their right-skewed nature. It also considers factors that may influence claim amounts. See \citet{frees2014predictive}.

Since the introduction of \citet{tweedie1984index} in the 1980s, the Tweedie mixture distribution has gained popularity in insurance claims modeling as it simplifies the modeling process by eliminating the need for separate frequency and severity models. The distribution itself exhibits a point mass at zero representing the probability of zero claims, followed by continuous density function for positive claim values. See \citet{frees2014predictive}. Such integration is a more accurate risk assessment as it also captures the inherent interdependence between the frequency and severity components. Primarily due to the distribution's exceptional flexibility to accommodate a diverse range of data, Tweedie regression models have a proven track record of providing superior fits to historical insurance claims data when compared to the conventional two-part models. It is true that the two-part frequency-severity models can accommodate ease of implementation of excess of zeros (in the frequency part) and can leverage modern machine learning to enhance prediction (in the severity component), e.g., \citet{shi2024leveraging}. However, this poses a significant drawback of being able to build a unified framework to allow us to deduce practical interpretation and understanding of the effect of the features affecting the aggregate loss, as we illustrate in the latter part of this paper.

With the influx of machine learning, the abovementioned conventional methods started to adopt powerful boosting algorithms. Gradient boosting has become immensely popular in the field of machine learning and is increasingly being adopted in insurance and actuarial science for many reasons, among which include predictive accuracy and robustness. This technique was initially introduced by \cite{friedman2001greedy} and is based around the idea of combining multiple weak learners, such as small decision trees, to create a single, strong, and robust predictive model.
Suppose we are given a training dataset $\mathcal{D} = \{(\bm{x}_i, y_i)\}^n_1$. Gradient boosting generates a sequence of functions\, $W_0, W_1,\ldots, W_T$, by minimizing the expected value of a specified loss function, $\ell(y_i,W_t)$. This loss function takes two inputs: the $i$-th response $y_i$ and the $t$-th function $W_t$, which estimates $y_i$. To improve on these estimates of $y_i$, we iteratively search for another function $W_{t+1} = W_t + w_{t+1}(\bm{x})$ that minimizes the expected value of the loss function, where $w_{t+1}$ is the weak learner at the $(t+1)$-th iteration. That is,

\begin{equation} \label{eq1}
w_{t+1}=\argmin_{w \in H} \mathbb{E} [\ell(y, W_{t+1})].
\end{equation}

Let $H$ be the set of candidate weak learners being evaluated, with the objective of selecting one to incorporate into the model. Based on the definition of $W_{t+1}$, we can express the expected value of the loss function $\ell$ in terms of $W_t$ and $w_{t+1}$ as follows:

\begin{equation}  \label{eq2}
\mathbb{E}[\ell(y, W_{t+1})]=\mathbb{E}[\ell(y, W_t + w_{t+1})].
\end{equation}

Our goal is to minimize the value of the loss function with respect to $y$ and $W_t$, while considering the additional factor $w_{t+1}$. Assuming that the loss function is continuous and differentiable, this process involves adjusting $W_t$ towards the direction where the loss function decreases most steeply. This direction corresponds to the rate of change of the loss function. Thus, by aligning $w_{t+1}$ with the direction where the gradient of the loss function with respect to $W_t$ declines most sharply, we approximate a $w_{t+1}$ that closely approaches the minimum of $\sum_{i=1}^{n} \ell(y_i, W_{t+1}(\bm{x}_i))$. The direction of steepest descent in this context is the negative gradient. Consequently, we define the pseudo-residuals $r_{i,t+1}$ as follows: 
\begin{equation} \label{eq:4}
r_{i,t+1}=-\frac{\partial \ell(y_i,W_t(\bm{x}_i))}{\partial W_t(\bm{x}_i)},
\end{equation}
and fit $w_{t+1}$ closely to these pseudo-residuals.

Since we are interested in modeling aggregate insurance losses, we design boosting algorithms to optimize loss functions on the distribution of the target variable, e.g., Tweedie loss function calculated as the negative log-likelihood of the data. Insurance claims data frequently features a significant number of zero values, especially for low-frequency events or claims. Additionally, the data often includes a small number of high-value claims, leading to a skewed distribution. Given these inherent characteristics of insurance claims data, choosing a loss function associated with the Tweedie distribution is a promising strategy when incorporating gradient boosting methods. In the case of the Tweedie loss function, it incorporates three parameters $\mu$, $p$, and $\phi$  that all control the shape of this function. We find from Figure 1(a) the lack of flexibility of these parameters to control the function in handling the zeros and right-skewness of the data.


Henceforth, we propose an enhancement to the Tweedie distribution through the introduction of an inflation probability, denoted as $q$. This modification leads to the formulation of what we refer to as the Zero-Inflated Tweedie distribution. In this revised distribution, the probability of zero is represented by the expression $(q + (1-q)\times f_{Tw}(y=0))$, where $f_{Tw}(y=0)$ denotes the probability of a zero from the original Tweedie distribution. In order to provide an even further flexibility in accommodating the inherent characteristics of the data, we suggest introducing the inflation probability $q$ as a function dependent on $\mu$, with an additional parameterization $\gamma$. Figure 1(b) showcases a more flexible $q$ as a function of $\mu$, addressing the limitations of the Tweedie distribution. 

\begin{figure}[!ht]
\centering
\begin{subfigure}{.48\textwidth}
\centering
\includegraphics[width=1.0\linewidth]{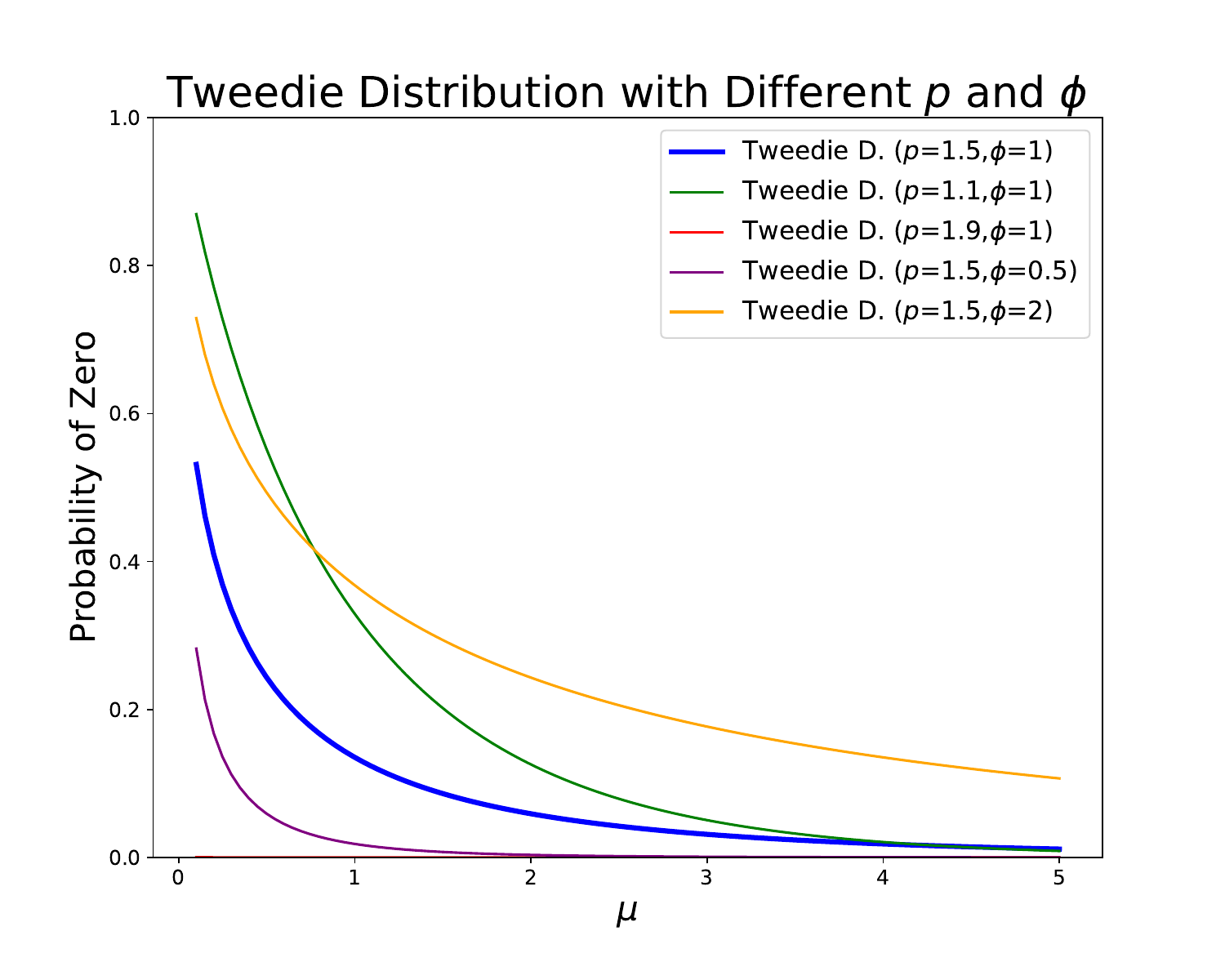}
\caption{Tweedie distribution}
\label{fig:combined_dist1}
\end{subfigure}
\begin{subfigure}{.48\textwidth}
\centering
\includegraphics[width= 1.0\linewidth]{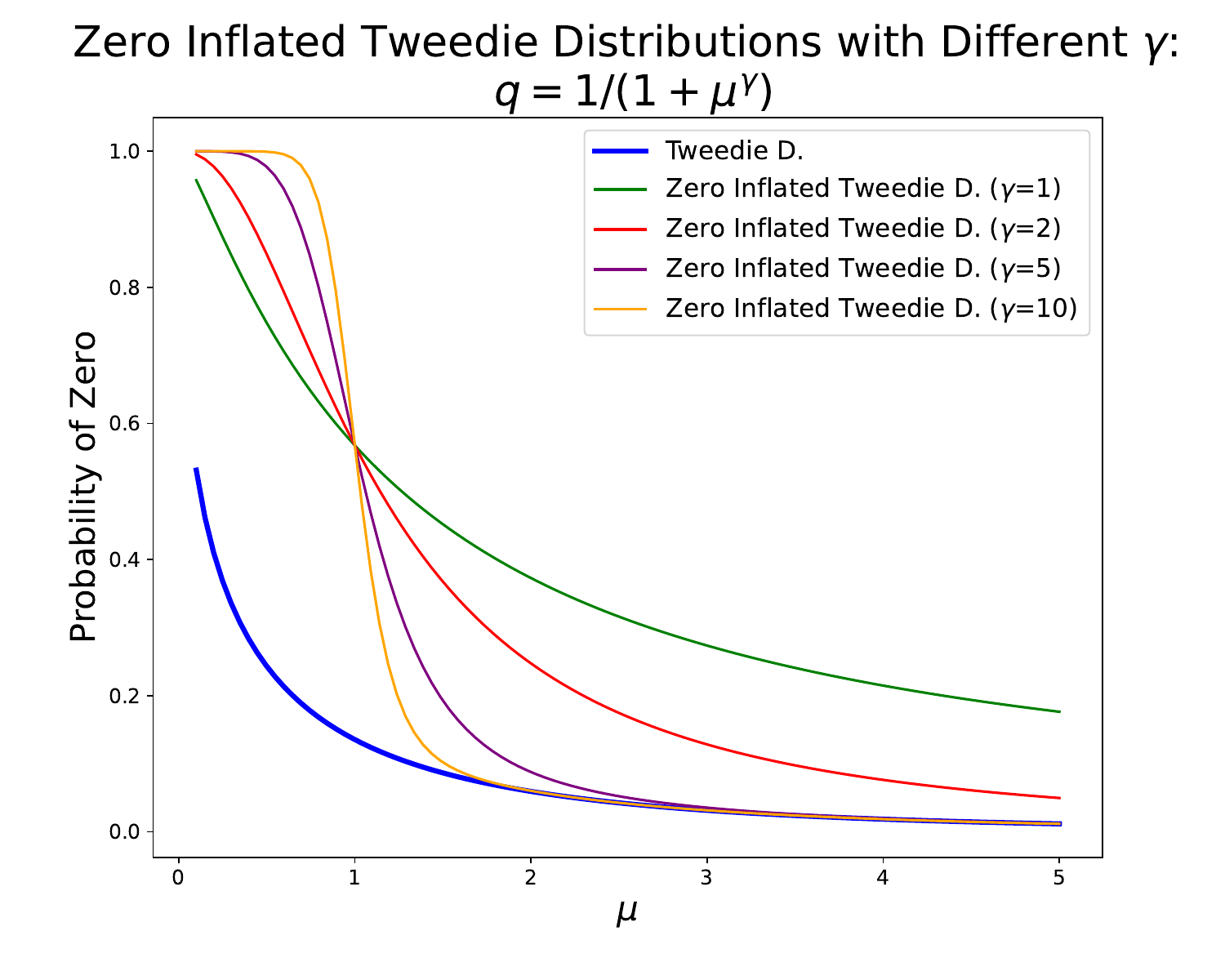}
\caption{Zero-Inflated Tweedie Distribution}
\label{fig:double_lift_bg_test}
\end{subfigure}
\caption{Comparing Tweedie and Zero-Inflated Tweedie Distributions}
\label{fig:combined_dist2}
\end{figure}

To address the challenges of zero-inflated, skewed data and to enhance the flexibility of modeling insurance losses, our paper makes several key contributions.
First, we implemented a customized loss function based on the zero-inflated Tweedie distribution described above. This decision was made because we believe the traditional Tweedie distribution is insufficient to provide a more accurate characterization of insurance losses. While there are some papers  that discuss its potential, the use of zero-inflated Tweedie distribution remains limited in the actuarial and statistical literature. Second, we reparameterized the zero-inflated Tweedie loss function to model the inflation probability $q$ as a function of the mean parameter $\mu$. This relationship between $q$ and $\mu$ results in a unified model that simplifies interpretation and improves computational efficiency. This approach is more appealing to practitioners, increasing the likelihood of its adoption in practice. In the data analysis section, we demonstrated various methods for interpreting the model's estimates and outcomes, which are made possible by our unified model. In addition, in contrast to the standard Tweedie model where frequency and severity are assumed independent, we allow for a direct link between $q$ and $\mu$. This implies some form of dependence or correlation between loss frequency and severity, which makes the model much more flexible as indicated by \citet{shi2020regression}. Third, for empirical demonstration, we analyzed a synthetic telematics dataset drawn from \citet{so2021synthetic}. This data contains records of 100,000 automobile insurance policies and showcases a notable zero-inflation pattern, with only 2,698 policies experiencing at least one claim. We specifically picked this dataset to allow us to examine the need for any specific allowance or special treatments related to compositional features. It is known that when GLM (Generalized Linear Model) is used, we need adjustments to the compositional feature variables to avoid bias. However, we found that such adjustments become unnecessary when boosting algorithms are employed. Finally, we implemented \texttt{CatBoost}, among other boosting libraries, to train our Gradient Boosted Decision Tree (GBDT) models. In \citet{so2024enhanced}, So explored various boosting libraries, including \texttt{XGBoost}, \texttt{LightGBM}, and \texttt{CatBoost}, on the same dataset for estimating various frequency models and found that \texttt{CatBoost} is the most efficient.  For our purposes, \texttt{CatBoost} further offers enhanced efficiency in handling special types of feature variables, particularly categorical features that are not uncommon in insurance data. By leveraging the zero-inflated Tweedie loss function and employing \texttt{CatBoost}, we were able to create a more flexible and accurate model for predicting insurance losses. These adjustments not only improved the model's performance but also enhanced its interpretability and ability to handle real-world insurance data effectively.

The structure of this paper is outlined as follows. Section \ref{sec:sc2} provides an overview of zero-inflated Tweedie boosted trees, discussing their relevance in modeling zero-inflated data commonly encountered in various domains. Section \ref{sec:catB} provides an in-depth discussion about \texttt{CatBoost}, an efficient gradient boosting algorithm known for its robustness and performance in handling categorical variables. Section \ref{sec:sc3} presents a comprehensive overview of the methodology employed in the study, including data preprocessing steps, model training techniques, and evaluation metrics used to assess model performance. Section \ref{sec:empirical} delves into the empirical analysis of the dataset, showcasing the application of the proposed methods (zero-inflated Tweedie boosted trees and \texttt{CatBoost}) using a synthesized data containing telematics information. We implemented several ways of interpreting the model estimates and outcomes. Section \ref{sec:conclude} concludes this paper. 

\section{Zero-Inflated Tweedie Boosted Decision Trees} \label{sec:sc2}

This section provides an overview for a comprehensive understanding of the Tweedie model, particularly focusing on its application in modeling the aggregate claim amount. It serves as a precursor to detailed discussions on the zero-inflated Tweedie model and the zero-inflated Tweedie boosted tree model. To begin, let us consider a Poisson random variable $N$, which represents the number of claims, and a sequence of independent and identically distributed (i.i.d.) gamma random variables $X_i$, for $i=1,2,\ldots,N$. By assuming independence between $N$ and the $X_i$'s, we can define the random variable $Y$ as follows:

\begin{equation} \label{eq:15}
Y=
\begin{cases}
	0, & \text{if } N=0,\\[1ex]
	X_1 + X_2 + \cdots + X_N, & \text{if } N = 1, 2, \ldots.
\end{cases}
\end{equation}

Here, $Y$ represents the Poisson sum of independent gamma random variables. In the context of actuarial science, we interpret $Y$ as the aggregate claim amount, where $N$ signifies the count of claims, and each $X_i$ represents the amount associated with the $i$-th claim. This specific distribution of $Y$ is known as the compound Poisson-gamma distribution.

\citet{smyth1996regression} highlighted that the compound Poisson-gamma distribution falls within a distinctive class of exponential dispersion distributions known as Tweedie distributions (\citet{tweedie1984index}). The Tweedie distributions belong to the linear exponential family, characterized by a disperson parameter $\phi$ and a power parameter $p$, with known mean and variance functions as follows:
\begin{itemize}
\item[] mean: $\mathbb{E}(Y) = \mu$
\item[] variance: $\text{Var}(Y) = \phi \mu^p$
\end{itemize} 
When constructing the compound Poisson-gamma model for the insurance aggregate claims, we restrict the power parameter $p$ to the interval $1 < p < 2$. The Tweedie models, denoted as $\text{Tw}(\mu, \phi, p)$, are defined by the following probability density function:
\begin{equation} \label{eq:16}
f_{\text{Tw}}(y|\mu, \phi, p)=a(y, \phi, p)\exp\left(\frac{1}{\phi}\left(y\frac{\mu^{1-p}}{1-p}-\frac{\mu^{2-p}}{2-p}\right)\right), \;\; y \geq 0,
\end{equation}
where \(a(\cdot)\) represents a normalizing function, \(\mu > 0\) denotes the expected value of \(Y\), and \(\phi > 0\) represents the dispersion parameter. Throughout this paper, we aim to delve deeper into the Tweedie model described above, with specific emphasis on expanding its applicability and understanding.

The boosted tree model is an approach that assumes the logarithm of the expected value of the target variable $Y$, given a set of features $\bm{x}$, can be effectively modeled using the prediction scores generated by gradient boosting. This relationship can be expressed as follows:
\begin{equation} \label{eq:17}
\ln \mathbb{E} (Y\mid \bm {x} ) = \ln E + W_T(\bm{x}), 
\end{equation}
where $\ln E$ is the offset term and $W_T(\bm{x})$ denotes 
the prediction score produced by summing the prediction of individual trees in the gradient boosting model, expressed as:
\[	W_T(\bm{x})=w_1(\bm{x})+w_2(\bm{x})+\cdots +w_t(\bm{x})+\cdots+w_T(\bm{x}).\]
Here, $w_t(\bm{x})$ represents the prediction of the $t$-th tree in the gradient boosting model. This framework allows for a flexible and powerful modeling of complex relationships between features and the target variable.

In the context of the Tweedie distribution, the Tweedie boosted tree extends this concept by assuming that the response variable $Y$ follows a Tweedie model. Thus, the parameter $\mu$ in the Tweedie model aligns with the expected value $\mathbb{E} (Y\mid \bm {x})$ in the boosted tree model. This integration of the Tweedie distribution enhances the model's ability to handle insurance-related data, which often exhibits characteristics such as zero-inflated and right-skewness.

While libraries like \texttt{CatBoost} offer built-in parameters for training Tweedie boosted trees, challenges can arise in scenarios where observed zeros in insurance claim data exceed that expected under the Tweedie model assumption. To overcome this challenge, we adopted a ``zero-inflated'' approach, known as the zero-inflated Tweedie (ZITw) model.

The ZITw model combines a point mass at zero with the Tweedie model, improving the accuracy of estimating $\mu$ especially when dealing with zero-claim observations. The probability density function in the ZITw model can be formulated as follows:
\begin{equation} \label{eq:18}
f_{\text{ZITw}}(y|\mu, \phi, p, q)=
\begin{cases}
	q+(1-q)\cdot\exp\left(-\displaystyle\frac{1}{\phi}\displaystyle\frac{\mu^{2-p}}{2-p}\right),  &\quad  \text{if } y=0\\[3ex]
	(1-q)\cdot a(y, \phi, p)\cdot\exp\left(\displaystyle\frac{1}{\phi}\left(y\cdot\frac{\mu^{1-p}}{1-p}-\frac{\mu^{2-p}}{2-p}\right)\right), &\quad \text{if } y > 0.\\ 
\end{cases}
\end{equation}

In this formulation, $q$ represents the inflation probability, indicating the degree of zero inflation. The expected value of $Y$ under the ZITw model is determined by both $q$ and $\mu$, given by $(1-q)\mu$. Hence, estimating both $\mu$ and $q$ accurately becomes crucial, and the gradient boosting framework provides techniques for achieving this, which we further discuss in Section \ref{sec:sc3}.

\section{Categorical Boosting: \texttt{CatBoost}} \label{sec:catB}

Boosting algorithms have gained immense popularity among actuaries as they are considered one of the most advanced learning algorithms developed in recent years. This is because boosting is an iterative fitting process that combines weak learners -- those slightly better than random -- into a strong learner, resulting in more precise and accurate predictions. This process is often referred to as a stage-wise additive model, where a new weak learner is added at a time while existing weak learners in the model remain fixed and unchanged.

The first boosting algorithm, \texttt{AdaBoost.M1}, was developed by \citet{freund1997decision}. Initially designed for classification tasks, it has since been extended to include regression problems (\citet{hastie2009}). \citet{friedman2001greedy} introduced gradient boosting, an algorithm tailored for regression, which constructs a predictive model by leveraging numerical optimization to minimize the model's loss through a gradient descent process. Gradient Boosting Decision Tree (GBDT) is a specific type of gradient boosting that employs decision trees as weak learners. Notable software libraries for GBDT implementation include \texttt{XGBoost}, \texttt{LightGBM}, and \texttt{CatBoost}.

\texttt{CatBoost}, developed by Yandex \citep{prokhorenkova2018catboost}, is recognized for its effectiveness in handling heterogeneous datasets, a common scenario in insurance data. Recent studies (\citet{so2024enhanced}) have demonstrated \texttt{CatBoost}'s superior performance compared to its counterparts when processing insurance data. This section will elaborate on \texttt{CatBoost}'s distinctive implementation strategy.

\texttt{CatBoost} is increasing in popularity for its effective implementation of the Gradient Boosting Decision Tree (GBDT) algorithm. It excels in handling heterogeneous data that includes both continuous and categorical features, while employing decision trees as weak learners. As explained in the preceding section, the process of the tree construction in \texttt{CatBoost} is sequential, with each new tree aiming to approximate the negative gradients $r$ of the loss function $\ell$ at the predictions made by the current function $W$. \texttt{CatBoost} provides a variety of built-in loss functions, such as Poisson and Tweedie, that can be utilized with GBDT. Additionally, users have the option to implement customized loss functions by writing functions that return both the gradient and the Hessian, which are then used to find a tree that minimizes the loss function. When incorporating a new tree $w$, \texttt{CatBoost} uses a score function to assess potential trees. For a prospective tree $w$, the score function is defined as follows:
\begin{equation} \label{eq:11}
S(w(\bm{x}_i), r_i) = -\sum_i (w(\bm{x}_i) - r_i)^2.
\end{equation}

The structure of the new tree is determined by the index $j$ of specific features and a cutoff value $c$. The objective is to determine the optimal values $j$ and $c$ that maximize the score function. Let $x_{i,j}$ denote the value of the $j$-th feature on the $i$-th instance. For a tree $w$ with a depth of one, for example, the function $w(\bm{x}_i)$ and the score function are given by:
\[w(\bm{x}_i)=
\begin{cases} 
a_{\text{left}}, & \text{if } x_{i,j}\leq c\\[1ex]
a_{\text{right}}, & \text{if } x_{i,j}> c,
\end{cases}\]
and
\begin{equation} \label{eq:12}
S(w(\bm{x}_i),r_i)=- \left(\sum_{i:x_{i,j}\leq c} (a_{\text{left}}-r_i)^2 +\sum_{i:x_{i,j}> c} (a_{\text{right}}-r_i)^2 \right).
\end{equation}

Given $i$ and $j$, the optimal values for $a_{\text{left}}$ and $a_{\text{right}}$ can be determined by employing a regularized loss function $L$. To avoid overfitting, a regularized loss function can be used to limit the complexity of the trees: the regularization term is added to the loss function and penalizes the model for having too many leaves or a tree that is too deep. At the $t$-th iteration, the regularized loss function is expressed as:
\begin{equation} \label{eq:6}
L=\sum_{i=1}^{n} \ell(y_i, W_t(\bm{x}_i)) +\sum_{j=1}^{t} u(w_j),
\end{equation}
where
\begin{equation} \label{eq:7}
	W_t(\bm{x}_i)=W_{t-1}(\bm{x}_i)+w_t(\bm{x}_i),
\end{equation}
and $u(w_j)$ serves as the regularization term, penalizing the complexity of the tree $w_j$.

At each iteration $t$, we train a single decision tree and integrate it into the existing model, as depicted in equation (\ref{eq:8}).
\begin{equation} \label{eq:8}
L=\sum_{i=1}^{n} \ell(y_i,W_{t-1}(\bm{x}_i)+w_t(\bm{x}_i)) + u(w_t) +\text{constant}.
\end{equation}

By expanding the regularized loss function up to the second order using Taylor series and eliminating constants, the regularized loss function at the $t$-th iteration can be simplified as:
\begin{equation} \label{eq:9}
L=\sum_{i=1}^{n} \left[ g_{i,t} w_t(\bm{x}_i) + \frac{1}{2}h_{i,t} w_t(\bm{x}_i)^2\right] +u(w_t),
\end{equation}
where $g_{i,t}=\partial_{W_{t-1}} \ell(y_i,W_{t-1}(\bm{x}_i))$ and $h_{i,t}=\partial^2_{W_{t-1}} \ell(y_i,W_{t-1}(\bm{x}_i))$.

When incorporating L2 regularization with parameter $\lambda$ and restructuring equation (\ref{eq:9}) by leaves, we obtain:
\begin{equation} \label{eq:13}
L=\sum_{i \in \text{left}}\left[ g_{i,t} a_{t, \text{left}} + \frac{1}{2}h_{i,t} a_{t, \text{left}}^2\right] +\frac{1}{2}\lambda a_{t, \text{left}}^2 + \sum_{i \in \text{right}}\left[ g_{i,t} a_{t, \text{right}} + \frac{1}{2}h_{i,t} a_{t, \text{right}}^2\right] +\frac{1}{2}\lambda a_{t, \text{right}}^2.
\end{equation}

We aim to find the optimal value for each leaf $a_{t, .}$ by minimizing the regularized loss function specified in equation (\ref{eq:13}). Therefore, by taking derivatives with respect to each leaf value $a_{t, .}$, we can identify the optimal values: 
\begin{equation} \label{eq:100}
a_{t, \text{left}}= -\frac{\sum_{i \in \text{left}} g_{i,t}}{\sum_{i \in \text{left}} h_{i,t} +\lambda} \qquad \text{and} \qquad a_{t, \text{right}}= -\frac{\sum_{i \in \text{right}} g_{i,t}}{\sum_{i \in \text{right}} h_{i,t} +\lambda}  \;\;.
\end{equation}

By combining equations (\ref{eq:100}) and (\ref{eq:12}), we can determine the optimal values for $j$ and $c$ that maximize the score function.

\texttt{Catboost} stands out among GBDT libraries due to its unique handling of categorical variable. Other libraries such as \texttt{XGBoost} and \texttt{LightGBM} do not directly support, but need some preprocessing, in handling categorical data. It is important to note that categorical features do not inherently possess a natural ordering, and therefore, before making a split in the decision tree, categorical features must be converted into numerical values. \texttt{CatBoost} addresses this challenge by employing a technique known as ``Ordered Target Statistic.'' This method involves randomly rearranging the dataset and then performing target encoding for each instance, taking into account only the instances that precede the current one. The process of transforming categorical features into their numerical counterparts in \texttt{CatBoost} involves the following steps:
\begin{itemize}
\item The training instances are permutated in random order.
\item The target (or response) variable is converted from a floating point to an integer, with the target values divided into buckets for regression problems.
\item Categorical features are encoded to numerical features.
\end{itemize}
After splitting target values into $K$ buckets, the target variable takes integer values in the range from $0$ to $K-1$. For an instance having target value $i$, the encoding of the categorical feature is expressed as:
\[\text{encoding}^i_u=\frac{\text{countInClass}+\text{prior}}{\text{totalCount}+1}\]
Here, $u$ denotes one of the classes for the categorical feature. The term ``countInClass" represents the frequency of instances with class \(u\) whose target values are equal to $i$. ``totalCount" is the total number of instances with class \(u\). The ``prior" parameter, which is predetermined, is added to the encoding formula to ensure smoother results.

It is essential to ensure that the target encoding is solely based on instances preceding the current one to prevent data leakage. Consequently, both ``countInClass" and ``totalCount" are computed following this rule. For further details, please see \citet{hancock2020catboost}. 
	

\section{Methodology} \label{sec:sc3}

In conventional zero-inflated models, training is usually conducted separately for the mean $\mu$ and the inflation probability $q$. This approach requires the zero-inflated Tweedie (ZITw) boosted tree to use twice the number of trees compared to the Tweedie (Tw) model because each parameter is independently estimated as shown below:
\begin{equation} \label{eq:19}
\ln \mu= \ln E+\displaystyle{W^{mean}_T(\bm{x})},
\end{equation}
\begin{equation} \label{eq:20}
\text{logit}(q)	=\ln\frac{q}{1-q} =  W^{prob}_T(\bm{x}). 
\end{equation}

From a practical standpoint, this leads to a final model consisting of two separate boosted trees, making it challenging to interpret feature effects and analyze feature interactions. To address this issue, we propose a novel algorithm that models $q$ as a function of $\mu$, which allows for the direct estimation of $q$ from $\mu$. This approached is supported by \citet{so2024enhanced}, who demonstrated that a zero-inflated Poisson boosted tree, treating $q$ as a function of $\mu$, outperformed traditional models when applied to auto telematics data. The subsequent sections describe the operational mechanics of the ZITw boosted tree in these contexts.

Moreover, we discuss strategies to address the specific challenges posed by compositional data. Compositional data is characterized by feature values that collectively sum to a constant value, typically 1. Typical examples of such feature variables, which are commonly found in auto telematics data, are the percentages of time driving for each day of the week. The importance of this discussion is highlighted by our analysis of auto telematics data using the proposed techniques. Auto telematics data serves as a prime example of compositional data, making it a relevant setting for our investigation.
 
\subsection{Scenario 1: Functionally unrelated} \label{sub:unrelated}

We refer to Scenario 1 as the methodology that pertains to the case where $q$ is not directly funtionally related to $\mu$. In this case, as indicated by equations (\ref{eq:19}) and (\ref{eq:20}), it becomes crucial to train $W^{mean}_T(\bm{x})$ and $W^{prob}_T(\bm{x})$ separately. This separation is necessary because $W^{mean}_T(\bm{x})$ represents the prediction score for the mean parameter $\mu$, which is obtained through an exponential transformation, while $W^{prob}_T(\bm{x})$ denotes the prediction score for the inflation probability $q$, which is achieved via a sigmoid transformation. The loss function used in this context is defined as follows:
\begin{equation} \label{eq:21}
\ell(y,W^{mean}_T(\bm{x}), W^{prob}_T(\bm{x})) =
\begin{cases}
-\ln \left(q+(1-q)\cdot\exp\left(-\displaystyle\frac{1}{\phi}\displaystyle\frac{\mu^{2-p}}{2-p}\right)\right), &\quad \text{if } y=0\\[3ex]
-\ln(1-q)-\ln a(y, \phi, p) - \left(\displaystyle\frac{1}{\phi}\left(y\cdot\frac{\mu^{1-p}}{1-p}-\frac{\mu^{2-p}}{2-p}\right)\right), &\quad \text{if } y>0,\\ 
\end{cases}
\end{equation}
where $\mu=E e^{W^{mean}_T(\bm{x})}$ and $q=\text{logit}^{-1}(W^{prob}_T(\bm{x}))$. The gradients and Hessians of the loss function with respect to $W^{mean}_t(\bm{x})$ are given by:
\begin{equation} \label{eq:22}
g^{mean}_{t+1}=\partial_{W^{mean}_{t}(\bm{x})} \ell(y,W^{mean}_{t}(\bm{x}),W^{prob}_{t}(\bm{x}))= \begin{cases}
\displaystyle\frac{\alpha\cdot \beta}{q+\alpha},&\quad \text{if } y=0\\[3ex]
-\displaystyle\frac{1}{\phi}y\mu^{1-p}+\beta,	&\quad \text{if } y>0,\\ 
\end{cases}
\end{equation}

\begin{equation} \label{eq:23}
h^{mean}_{t+1}=\partial^2_{W^{mean}_{t}(\bm{x})} \ell(y,W^{mean}_{t}(\bm{x}),W^{prob}_{t}(\bm{x}))=
\begin{cases}
\displaystyle\frac{\alpha\cdot \beta((2-p-\beta)\cdot(q+\alpha)+\alpha\cdot \beta)}{(q+\alpha)^2}, &\quad \text{if } y=0\\[3ex]
-\displaystyle\frac{1}{\phi}y(1-p)\mu^{1-p}+(2-p)\cdot \beta,	    &\quad \text{if } y>0,\\ 
\end{cases}
\end{equation}
where $\alpha=(1-q)\exp\left(-\displaystyle\frac{1}{\phi}\displaystyle\frac{\mu^{2-p}}{2-p}\right)$, and $\beta=\displaystyle\frac{1}{\phi}\mu^{2-p}$.

Similarly, the gradients and the Hessians of the loss function with respect to $W^{prob}_t(\bm{x})$ are given by: 
\begin{equation} \label{eq:24}
g^{prob}_{t+1}=\partial_{W^{prob}_{t}(\bm{x})} \ell(y,W^{mean}_{t+1}(\bm{x}),W^{prob}_{t}(\bm{x}))= \begin{cases}
\displaystyle\frac{1}{1+e^{\delta+W^{prob}_{t}(\bm{x})}}-\frac{1}{1+e^{W^{prob}_{t}(\bm{x})}},&\quad \text{if } y=0\\[3ex]
\displaystyle\frac{e^{W^{prob}_{t}(\bm{x})}}{1+e^{W^{prob}_{t}(\bm{x})}},&\quad \text{if } y>0,\\ 
\end{cases}
\end{equation}

\begin{equation} \label{eq:25}
h^{prob}_{t+1}=\partial^2_{W^{prob}_{t}(\bm{x})} \ell(y,W^{mean}_{t+1}(\bm{x}),W^{prob}_{t}(\bm{x}))=\begin{cases}
\displaystyle\frac{e^{W^{prob}_{t}(\bm{x})}}{\left(1+e^{W^{prob}_{t}(\bm{x})}\right)^2}-\frac{e^{\delta+W^{prob}_{t}(\bm{x})}}{\left(1+e^{\delta+W^{prob}_{t}(\bm{x})}\right)^2},&\quad \text{if } y=0\\[3ex]
\displaystyle\frac{e^{W^{prob}_{t}(\bm{x})}}{\left(1+e^{W^{prob}_{t}(\bm{x})}\right)^2},&\quad \text{if } y>0,\\ 
\end{cases}
\end{equation}
where $\mu=E e^{W^{mean}_{t+1}(\bm{x})}$ and $ \delta=\displaystyle\frac{1}{\phi}\displaystyle\frac{\mu^{2-p}}{2-p}$.

To minimize the loss function with two separate models, we adopt the training technique proposed by \citet{meng2022actuarial}. This approach employs the principle of coordinate descent optimization, where during each iteration, one parameter (e.g., $\mu$ or $q$) is optimized while keeping the other fixed. This process is repeated alternately for $\mu$ and $q$ until convergence, leading to improve predictive accuracy for both parameters. More specifically, during the \( t \)-th iteration, with a fixed inflation parameter \( q \), we train \( w_t^{mean}(\bm{x}) \) (the \( t \)-th decision tree) to estimate the mean parameter \( \mu \). Then, with the estimated \( \mu \) held constant, we train \( w_t^{prob}(\bm{x}) \) to estimate the inflation probability \( q \). After completion of an estimation cycle for both \( \mu\) and \(q\), the dispersion parameter \( \phi\) is estimated by minimizing the mean deviation described in subsection \ref{subsub:metric1}. The power parameter is set to a predetermined constant, typically 1.5. For a detailed explanation of this training algorithm, please refer to Algorithm \ref{alg:case2} in Appendix A.

\subsection{Scenario 2: Functionally related} \label{sub:related}
We introduce Scenario 2 as a novel methodology wherein the parameter $q$ is functionally linked to $\mu$. This innovative approach, absent from existing literature, acknowledges the relationship between the inflation parameter $q$ and the mean parameter $\mu$. Given our limited prior understanding of this relationship, we propose a refinement, within the boosting algorithm, of the parameterization initially introduced by \citet{lambert1992zero} for zero-inflated Poisson Generalized Linear Models (GLM).

Inspired by this work, our proposed parameterization allows us to relate the canonical link functions of the zero component, which is the $\text{logit}(q)$, and the continuous component, which is the $\ln \mu$, as depicted in the following equations:
\[	\ln \mu = \ln E+ W_T(\bm{x}), \]
\begin{equation} \label{eq:27}
\text{logit}(q)	=\ln\frac{q}{1-q} = -\gamma(\ln E+ W_T(\bm{x})). 
\end{equation}
From \eqref{eq:27}, we are able to derive the relationship between $q$ and $\mu$:
\begin{equation} \label{eq:26}
q=\frac{1}{1+\mu^{\gamma}} \;.
\end{equation}

The parameter $\gamma$ is estimated from the data. Equation (\ref{eq:26}) reveals that for $\gamma >0$, the inflation probability diminishes as $\mu$ increases. Guided by this assumption, we define the loss function as follows:
\begin{equation} \label{eq:28}
\ell (y, W_T(\bm{x})) =\begin{cases}
-\ln\left(1+\mu^{\gamma}\cdot\exp\left(-\displaystyle\frac{1}{\phi}\displaystyle\frac{\mu^{2-p}}{2-p}\right)\right)+\ln(1+\mu^{\gamma}), &\quad y=0\\[3ex]
 -\gamma \ln \mu +\ln(1+\mu^{\gamma})-\ln a(y, \phi, p)-\displaystyle\frac{1}{\phi}\left(y\cdot\frac{\mu^{1-p}}{1-p}-\frac{\mu^{2-p}}{2-p}\right), &\quad y>0,\\ 
\end{cases}
\end{equation}
where $\mu=Ee^{W_T(\bm{x})}$. The gradient and Hessian of the loss function with respect to $W_t(\bm{x})$ are:
\begin{equation} \label{eq:29}
g_{t+1}=\partial_{W_{t}(\bm{x})} \ell(y,W_{t}(\bm{x}))= \begin{cases}
-\displaystyle\frac{\zeta\cdot\eta}{1+\zeta}+\kappa,  &\quad y=0\\[3ex]
 -\gamma +\kappa -\displaystyle\frac{1}{\phi}y \mu^{1-p} +\displaystyle\frac{1}{\phi} \mu^{2-p}, &\quad y>0,\\ 
\end{cases}
\end{equation}

\begin{equation} \label{eq:30}
h_{t+1}=\partial^2_{W_{t}(\bm{x})} \ell(y,W_{t}(\bm{x}))=\begin{cases}
\displaystyle\frac{\left(\displaystyle\frac{2-p}{\phi}\mu^{2-p} -\eta^2\right)\cdot\zeta\cdot(1+\zeta)+ \zeta^2\cdot \eta^2 }{(1+\zeta)^2}+\kappa^2 \cdot \mu^{-\gamma},       &\quad y=0\\[3ex]
\kappa^2 \cdot \mu^{-\gamma}-\displaystyle\frac{(1-p)y}{\phi}\mu^{1-p}+\displaystyle\frac{(2-p)}{\phi}\mu^{2-p},  &\quad y>0,\\ 
\end{cases}
\end{equation}
where $\zeta=\mu^\gamma\cdot\exp\left(-\displaystyle\frac{1}{\phi}\frac{\mu^{2-p}}{2-p} \right)$, $\eta=\gamma-\displaystyle\frac{1}{\phi}\mu^{2-p}$, and $\kappa=\displaystyle\frac{\gamma\mu^{\gamma}}{1+\mu^{\gamma}}$.

After constructing an additional tree, we update the value of \( \mu \), determine the inflation parameter \( q \), and subsequently, determine the dispersion parameter \( \phi \)  by minimizing the mean deviance with the updated values of \( \mu \) and \( q \). Finally, with \( \mu \), \( q \) and \( \phi \) fixed, we estimate \( \gamma \) by minimizing the mean deviance. The power parameter remains consistently set to a predetermined constant, typically at 1.5. For a detailed description of the training algorithm, please refer to Algorithm \ref{alg:case1} in Appendix A.

\subsection{Adjustment of Compositional Data} \label{sub:compositional}
Compositional data is characterized by multiple non-negative features that sum up to a constant, typically 100\% or 1. Due to the inherent statistical dependence among these features, transformations are often necessary to map the data onto the real Euclidean space. This transformation facilitates the application of traditional statistical methodologies (\citet{aitchison1982comp}). Among the notable transformations are the logratio methods, which include the centered logratio transformation (CLR), the additive logratio transformation (ALR), and the isometric logratio transformation (ILR). See also \citet{aitchison1994principles}.

When dealing with compositional data comprising \( J \) features, denoted as \( \{ \bm{x}_{\cdot1}, \bm{x}_{\cdot2}, \ldots, \bm{x}_{\cdot J} \} \), where the features sum to 1, we refer to these features as a \( J \)-part composition. The CLR involves taking the logarithm of a part, standardized relative to the geometric mean of the logarithms of all parts. Mathematically, $\text{CLR}(j)$ is defined as:
\[
\text{CLR}(j) = \ln \left( \frac{\bm{x}_{\cdot j}}{\left( \prod_{i} \bm{x}_{\cdot i} \right)^{1/J}} \right), \quad j = 1, 2, \ldots, J.
\]
On the other hand, the ALR contrasts a specific reference part with all other parts. By selecting the \( d \)-th part as the reference, the ALR transformation is represented as:
\[
\text{ALR}(j|d) = \ln \left( \frac{\bm{x}_{\cdot j}}{\bm{x}_{\cdot d}} \right), \quad j \neq d.
\]
The ILR transformation, as proposed by \citet{egozcue2003isometric}, is expressed as:
\[
\text{ILR}(\bm{x}) = R \cdot \text{CLR}(\bm{x}),
\]
where \( \bm{x} \) is a \( J \times n \) data matrix comprising \( J \) features, and \( R \) is a \( (J-1) \times J \) matrix satisfying the condition:
\[
R R^T = I_{J-1}.
\]
The matrix $R$ is referred to as the contrast matrix (\citet{pawlowsky2015modeling}). Although these logratio transformation methods are equivalent up to linear transformations, the ALR does not preserve distances between data points in the original space. While CLR retains such distances, it can lead to a determinant of zero. On the contrary, the ILR overcomes these issues but results in transformed features that are complex combinations of the original features, making interpretation additionally challenging.

The purpose of logratio transformations is to eliminate the inherent spurious correlations among compositional features. Following transformation, dimensionality reduction and decorrelation of these features become feasible. However, researchers have explored applying exponential family principal component analysis (EPCA) to compositional data (\citet{gan2021compositional}, \citet{yin2024flexible}, \citet{avalos2018representation}). To apply EPCA to compositional data, it is initially transformed into the real Euclidean space via a logratio transformation, such as CLR. Probabilistic principal components analysis (PPCA) is a subset of EPCA (\citet{tipping1999probabilistic}). \citet{yin2024flexible} emphasized the superiority of PPCA in compositional data analysis, highlighting its ability to restore original features, which is not achievable with EPCA alone. Therefore, we confine our study by employing PPCA subsequent to CLR transformation for its analysis.

\section{Empirical Investigation} \label{sec:empirical}

In this section, we undertake a comprehensive comparative analysis to evaluate the effectiveness of the models proposed in Section \ref{sec:sc3} compared to conventional models like the Tweedie boosted tree and Tweedie generalized linear model (GLM). The primary objective of this analysis is to determine the most suitable aggregate claims models for managing auto claim datasets with excessive zeros. Additionally, we aim to explore whether the performance of the boosted model is influenced by the additional complexity introduced by the compositional features in the dataset; this helps us assess the necessity for specific treatments related to compositional features.

Our empirical analysis is based on a synthetic telematics dataset developed by \citet{so2021synthetic}. This dataset comprises of 100,000 policies and demonstrates a zero-inflation characteristic, with only 2,698 policies experiencing at least one claim. For this study, a total of eight different models were trained:
\begin{enumerate}
\item Zero-inflated Tweedie boosted tree with scenario 1 (ZITwBT1)
\item Zero-inflated Tweedie boosted tree with scenario 2 (ZITwBT2)
\item Tweedie boosted tree (TwBT)
\item Tweedie GLM (TwGLM)
\item ALR
\item CLR
\item ILR
\item PPCA after CLR transformation
\end{enumerate}
For the latter four models listed above, we applied the suitable logratio transformations only to zero-inflated Tweedie boosted tree with scenario 2 to avoid overwhelming the reader.

The Tweedie boosted tree (TwBT) and Tweedie GLM (TwGLM) models were implemented using the \texttt{CatBoost} and \texttt{statsmodels} Python packages, respectively. These models utilized the built-in loss function during training. On the other hand, ZITwBT1 and ZITwBT2 were developed using \texttt{CatBoost} with a customized loss function, as outlined in Algorithms 1 and 2 in Appendix A, to apply the specified training methods.

For the boosted trees in our study, we carefully set hyperparameters to ensure robust and comprehensive model training. Specifically, we established the number of trained trees, denoted as \( T \), to be 500. This choice was made to strike a balance between model complexity and computational efficiency. Next, we tackled the crucial hyperparameters of learning rate \( \alpha \) and L2 penalty parameter \(\lambda\). The learning rate, which controls the contribution of each tree to the overall model, was selected from a grid of values including 0.01, 0.05, and 0.10. This range allowed us to explore different rates of learning while avoiding overly aggressive or sluggish updates. Similarly, the L2 penalty parameter, influencing the regularization of our model, was chosen from a grid spanning values from 0 to 100 with intervals of 10. This grid search enabled us to strike a balance between model complexity and regularization strength, preventing overfitting and improving generalization.

The ZITwBT1 model required a different approach due to its unique structure. To train this model effectively, we divided the total number of trained decision trees \( T \) equally between two distinct boostings -- one for \( \mu \) and another for $q$. This division ensures fairness in comparison with other models and in alignment with total iterations across all models. Each boosting in the ZITwBT1 model underwent \( T=250 \) iterations, maintaining consistency in training depth and performance evaluation.

Throughout our experiments, we maintained a fixed maximum tree depth of 10 and a power parameter set consistently at 1.5 across all models. These choices were made based on domain knowledge and empirical observations to ensure stable and reliable model performance. Moving forward, we conducted rigorous model comparisons based on four key metrics, elaborated in the subsequent subsection. To ensure unbiased evaluations, we held 20\% of the total dataset as the test set, while leveraging a 3-fold cross-validation method on the remaining training dataset to determine optimal hyperparameters. This approach allowed us to assess model performance accurately and make informed decisions regarding model selection and refinement.

\subsection{Performance Evaluation} \label{sub:metric}

Comparing Tweedie models with zero-inflated Tweedie models presents a complex challenge because of their distinct assumptions and structural differences. This study examines a comprehensive evaluation and comparison of these models based on a diverse set of assessment metrics. Among these metrics are traditional measures like deviance and Mean Absolute Deviation (MAD), which assess the overall goodness of fit and prediction accuracy. The study also utilizes the Vuong test, a statistical tool specifically designed for comparing non-nested models and determining their relative performance. Furthermore, the analysis includes two variants of the Gini index, a widely-used metric for evaluating predictive power and discrimination ability in models. By employing this array of assessment metrics, this paper aims to provide a rigorous comparison between Tweedie and zero-inflated Tweedie models, which can provide insights on their respective strengths and limitations in modeling insurance claim data and other similar applications.

\subsubsection{Deviance and Mean Absolute Deviation}\label{subsub:metric1}
Deviance quantifies the difference between a statistical model and a saturated model, which perfectly predicts the data. This is done by comparing the log-likelihoods of the two models and is a key concept in GLMs. Although zero-inflated boosted models are not exactly GLMs, \citet{martin2016r} investigated whether deviance could be similarly defined for zero-inflated models. Their research, which included simulations and real-world examples, produced positive results which suggest that deviance can be defined for zero-inflated models.

The unit deviance for the $i$-th observation in zero-inflated Tweedie models can be determined by the equation below:

\begin{equation} \label{eq:50}
d(y_i,(\hat{\mu}_i, \hat{\phi}, p, \hat{q}_i))=\begin{cases} 
-2\ln\left(\hat{q}_i + (1 -\hat{q}_i) \cdot \exp \left( -\displaystyle\frac{\hat{\mu}_i^{2-p}}{\hat{\phi}(2-p)} \right) \right), & \text{if } y_i = 0 \\
2\left( \displaystyle\frac{1}{\hat{\phi}} \left( \displaystyle\frac{y^{2 -p}}{1 -p} - \frac{y^{2 -p}}{2 -p} \right)
- \ln(1 - \hat{q}_i)-\displaystyle\frac{1}{\hat{\phi}} \left( \displaystyle\frac{y_i \cdot \hat{\mu}_i^{1 -p}}{1 -p} - \frac{\hat{\mu}_i^{2 -p}}{2 -p} \right)\right), &\text{if } y_i>0.
\end{cases}
\end{equation}
The mean deviance represents the average of the unit deviances for a specific model across all observations. A significant decrease in deviance suggests a better alignment between the model and the data.

The mean absolute deviation (MAD) is a statistical measure used to quantify the average absolute difference between the observed values and the predicted values, defined as: 
\[\text{MAD}=\frac{1}{n}\sum_{i=1}^n |y_i-\hat{y}_i|.\]
The MAD offers a simple way to assess the error in the predictions. A lower MAD value suggests higher precision in the predictions because it indicates that, on average, the predicted values are closer to the actual values.

\subsubsection{Vuong Test}\label{subsub:metric2}

The Vuong test, proposed by \citet{vuong1989likelihood}, provides a reliable method for comparing the zero-inflated Tweedie model with other non-nested models, such as the Tweedie model. Let $f_K(y_i \bm{x}_i)$ be the probability distribution functions for the i-th observation from model $K$. When both models exhibit similar data fit, their respective likelihood functions tend to be nearly identical. Conversely, any disparities in the likelihood functions can indicate which model provides a better fit. The Vuong test operates on this principle. We define $m_i$ as:
\[m_i=\ln \displaystyle\frac{f_1(y_i|\bm{x}_i)}{f_2(y_i|\bm{x}_i)}\]
We will test the hypothesis that the expected value of $m_i$ equals zero, with the Vuong statistic to gauge this:
\[V=\displaystyle\frac{\sqrt{n}\left(\displaystyle\frac{1}{n}\sum_{i=1}^{n}m_i \right) }{\sqrt{\displaystyle\frac{1}{n}\sum_{i=1}^{n}(m_i-\bar{m})^2}}\]
Under the assumption of the null hypothesis, the Vuong statistic follows an asymptotic normal distribution. If the statistic surpasses 1.96, it indicates that the first model is more likely to be correct; conversely, if it falls below -1.96, the second model is deemed more accurate, at a 5\% significance level.

\subsubsection{Gini Index ($\text{Gini}^a$ and $\text{Gini}^b$)} \label{subsub:metric3}
The Gini index is a well-established metric for assessing model prediction performance. This paper will specifically examine two popular variants of the Gini index, $\text{Gini}^a$ and $\text{Gini}^b$, frequently used in insurance modeling.

\noindent\textbf{$\text{Gini}^a$}  It has the following expression:

\[\text{Gini}^a=\displaystyle\frac{\displaystyle\frac{\sum_{i=1}^{n}y_iR(\hat{y}_i)}{\sum_{i=1}^{n} y_i}-\sum_{i=1}^{n}\frac{n-i+1}{n}}{\displaystyle\frac{\sum_{i=1}^{n}y_iR(y_i)}{\sum_{i=1}^{n} y_i}-\sum_{i=1}^{n}\frac{n-i+1}{n}}\]
In this context, \(R(y_i)\) represents the rank of \(y_i\). The \(\text{Gini}^a\) metric depends on the model's prediction of the expected claim amount. A higher \(\text{Gini}^a\) value indicates an improved discriminatory power of the model.

\noindent\textbf{$\text{Gini}^b$} The \(\text{Gini}^b\) metric, as proposed by \citet{frees2014insurance}, offers a sophisticated approach for comparing a collection of models. 

\noindent Consider two predictive aggregate claim amounts: 
	\begin{itemize}
		\item \(\hat{y}^{B}_i\) from the reference model, and 
		\item \(\hat{y}^{P}_i\) from the model under comparison.
	\end{itemize}
Their relative difference can be represented as:
	\[R_i = \frac{\hat{y}^{P}_i}{\hat{y}^{B}_i}.\]
Subsequent to determining \(R_i\), the instances are sorted in ascending order based on its values. The ordered Lorenz curve can then be drawn as:
	\[\left(\frac{\sum_{i=1}^{n}\hat{y}^{B}_i \cdot I[R_i\leq s]}{\sum_{i=1}^{n}\hat{y}^{B}_i}, \frac{\sum_{i=1}^{n}y_i \cdot I[R_i\leq s]}{\sum_{i=1}^{n} y_i}\right) \quad \text{for } s \in [0,1],\]
where \(I[\cdot]\) is the indicator function.

The Gini$^b$ coefficient is computed by doubling the area between the ordered Lorenz curve and the line \(y=x\). When evaluating multiple models, the one with the highest Gini$^b$ coefficient is deemed the most suitable compared to a reference model. To determine the best model, each model is considered as the base model in turn, and the maximum Gini$^b$ value for each base model is recorded. The model with the lowest Gini$^b$ among these maximum values is then selected as the optimal model. This selection method, known as the ``min-max'' approach, helps identify and discriminate the most dependable model among the alternatives (\cite{frees2014insurance}).

\subsection{Synthetic Telematics Data} \label{sub:data2}

\begin{figure}[htbp]
\centering
\includegraphics[scale=0.8]{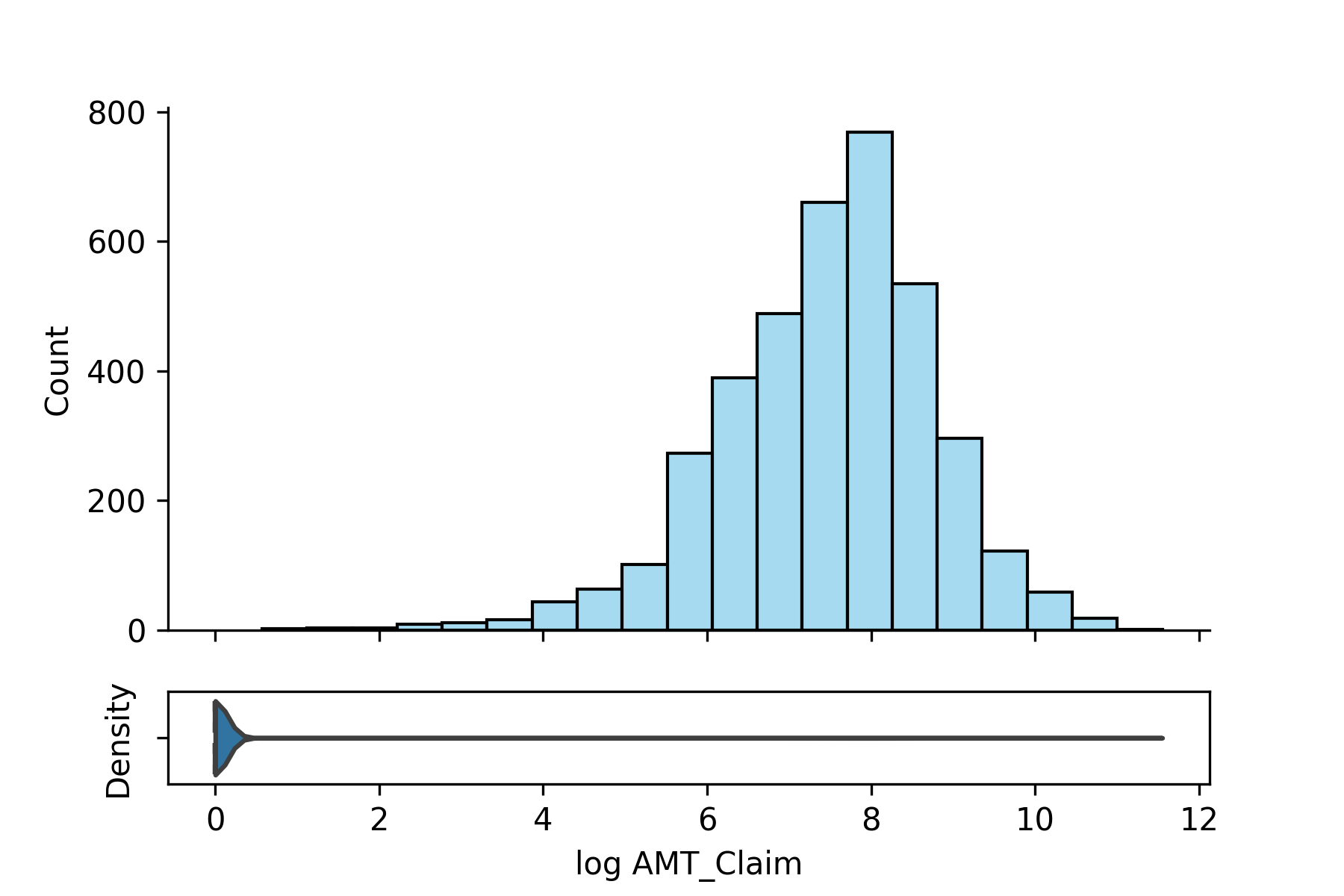}
\caption{Distribution of Aggregate Claim Amounts, on a log scale, for the Synthetic Telematics Data}	\label{fig8:hist_syn}
\end{figure}

The synthetic telematics dataset developed by \citet{so2021synthetic} used in this study is publicly accessible at \url{http://www2.math.uconn.edu/~valdez/data.html}. This dataset was meticulously generated using Synthetic Minority Over-sampling Technique (SMOTE), coupled with a feedforward Neural Network (NN), and trained based on authentic data from a Canadian insurance provider. It consists of 100,000 data samples, with only 2,698 policies reporting at least one claim. This notable occurrence of zero-inflation within the dataset highlights a distinct characteristic of the data. The dataset is structured around 52 distinct features categorized into Traditional data (e.g., exposure, insured age, and gender), Telematics data (e.g., total miles driven, harsh braking, and harsh acceleration), and Claims Data (encompassing claim counts and aggregate claim amounts). Evidently, the aggregate claim amount models were rigorously developed utilizing exposure and 49 features, with the response variable being the claim amounts. Among these 49 features, five are categorical variables, each having varying numbers of categories with 2, 2, 4, 2, and 55 categories, respectively. A noteworthy attribute of this dataset is the presence of a few categorical attributes. This suggests the need for the selection of models capable of effectively handling such features to ensure optimal results and accurate data fitting. To this end, the \texttt{CatBoost} library was employed to implement boosted tree algorithms, which offers the necessary capabilities to handle these intricate features. For a detailed overview of the dataset and its attributes, readers are directed to Table \ref{tab:VD2} in Appendix B, where a comprehensive description of the variables in the dataset can be found. Figure \ref{fig8:hist_syn} provides an overview of the distribution of the aggregate claim amounts for the synthetic telematics data. The graph shows a point mass at zero and a histogram of the positive aggregate claim amounts. For readers who want to know more about this data, please see \citet{so2021synthetic}.

\subsubsection{Results of Model Performance}\label{subsub:metric4}

First, we examine how the zero-inflated Tweedie boosted tree models perform relative to the conventional Tweedie GLM and Tweedie boosted tree models ignoring the compositional features in the dataset. Table \ref{tab:metric} shows that the zero-inflated Tweedie boosted tree models (ZITwBTs) have similar MAD values, which are notably lower than those of the Tweedie Generalized Linear Model (TwGLM) and Tweedie boosted tree model (TwBT). This suggests that the zero-inflated models provide more accurate predictions. Furthermore, the deviances observed in the ZITwBT models are consistently lower compared to the TwGLM and TwBT models, which suggests the suitability of the zero-inflated approach for modeling our synthetic telematics data. In terms of \(\text{Gini}^{a}\), ZITwBT2 has a significantly higher value than the TwGLM, TwBT models, and the model from scenario 1 (ZITwBT1), which indicates that it performs better in training our synthetic telematics data. Therefore, Table \ref{tab:metric} suggests the strong promise of the use of the ZITwBT2 model as it outperforms these other models in training telematics data.

\begin{table}[htbp]
\caption{Comparison of MAD, Deviance, and $\text{Gini}^{a}$ across 4 Models}
\centering
\scalebox{1.10}{
	\begin{tabular}{l|c|c|c|c}
	\toprule
	 &TwGLM&TwBT&ZITwBT1&ZITwBT2\\ \hline
MAD&257&193&\color{red} 145& 146\\ \hline
Deviance&0.065&0.062&0.0018&\color{red}0.0017\\ \hline
$\text{Gini}^{a}$&0.634&0.736&0.194&\color{red}0.748\\ 
	 \bottomrule
\end{tabular}} \label{tab:metric}
\end{table}

Next, we investigate the impact of the compositional features on model performance, specifically focusing on ZITwBT2 models. Table \ref{tab:metric2} presents the outcomes of these models, comparing their performance with and without adjustments made to compositional features. The ZITwBT2 model, when implemented without any compositional feature adjustments, exhibits noteworthy results, as evidenced by the highest observed \(\text{Gini}^{a}\) value alongside the lowest MAD and deviance scores. This particular configuration emerges as the most reliable among the 8 models scrutinized, considering these three critical evaluation metrics. This preference of the ZITwBT2 model without compositional feature adjustments underscores its possible robustness in capturing the underlying patterns and nuances present in the dataset. By leveraging compositional features in their raw form, the model leads to more accurate predictions and a better understanding of the underlying data dynamics, which sheds light on how to handle compositional features.

\begin{table}[htbp]
\caption{Comparison of MAD, Deviance, and $\text{Gini}^{a}$ in ZITwBT2 Models with and without Compositional Data Adjustment}
\centering
\scalebox{1.10}{
	\begin{tabular}{l|c|c|c|c|c}
	\toprule
&\multirow{2}{*}{ZITwBT2}&\multicolumn{4}{c}{ZITwBT2}\\ \cline{3-6} 
&&alr&clr&ilr&clr+PPCA \\ \hline
MAD&\color{red}146&149&176&149&149\\ \hline
Deviance&\color{red}0.0017&0.0020&0.0026&0.0032&0.0025 \\ \hline
$\text{Gini}^{a}$&\color{red}0.748&0.740&0.712&\color{red}0.748&0.736 \\ \bottomrule
\end{tabular}} \label{tab:metric2}
\end{table}

Table \ref{tab:vuong} presents the Vuong statistics calculated for the four distinct models. This statistical metric, which serves as a comparative tool, allows us to assess the relative performance of these models. Specifically, when the Vuong statistic exceeds 2.33 at a 1\% significance level, it signifies that the first model is superior to the second. Conversely, a value lower than -2.33 indicates that the second model outperforms the first. The corresponding $p$-values, denoted in parentheses within the table, further support these comparisons. Upon examining the results, a notable trend emerges when considering the ZITwBT2 model as the first model and TwGLM and TwBT as the second models. In this scenario, the $p$-values approach zero, signifying a high level of statistical significance, while the Vuong statistic is positive. These combined indicators strongly suggest that the ZITwBT2 model exhibits superior performance compared to both TwGLM and TwBT. Furthermore, when juxtaposing the ZITwBT2 model as the first model against ZITwBT1 as the second model, a different pattern emerges. Here, the $p$-values surpass the 1\% threshold, indicating a lack of statistical significance in differentiating between the performances of these models. In essence, the evidence suggests that the performances of ZITwBT2 and ZITwBT1 are comparable, without a clear distinction in superiority at the specified significance level.

\begin{table}[htbp]
\caption{Vuong Statistics and Corresponding $p$-values across 4 models}\label{tab:vuong}
\centering
\begin{threeparttable}
\scalebox{1.15}{
\begin{tabular}{c|c|c|c|c|c|}
\multicolumn{2}{c}{}&\multicolumn{4}{c}{Second Model} \\ \cline{2-6}  &&TwGLM&TwBT&ZITwBT1&ZITwBT2\\ \cline{2-6} 
\multirow{4}{*}{\rotatebox[origin=c]{90}{First Model}}&TwGLM&&-&-&-\\ \cline{2-6}
&TwBT&0.93 (\color{red}0.35)&&-&-\\ \cline{2-6}
&ZITwBT1&27.56 (0.00)&17.03 (0.00)&&- \\ \cline{2-6}
&ZITwBT2&27.65 (0.00)&17.16 (0.00)&0.08 (\color{red}0.94)& \\ \cline{2-6}
\end{tabular}} 
\begin{tablenotes}
\scriptsize
\item Numbers represent Vuong statistics, with p-values shown in parentheses.	
\end{tablenotes}
\end{threeparttable}
\end{table}

The results depicted in Table \ref{tab:vuong2} provide further intriguing finding - both ZITwBT2 models, with and without compositional data adjustment, exhibit equivalent performance levels. This equivalence suggests that the inclusion or exclusion of compositional data adjustment does not significantly impact the predictive power of accuracy of the models. This analysis based on Vuong statistics indicates that the performance of ZITwBT2 models, both with and without compositional data adjustment, is equivalent. This equivalence further underscores the stability, consistency, and robustness of the ZITwBT2 models as reliable predictive models.

\begin{table}[htbp]
\caption{Vuong Statistics and Corresponding $p$-values in ZITwBT2 Models with and without Compositional Data Adjustment}\label{tab:vuong2}
\centering
\begin{threeparttable}
\scalebox{1.10}{
\begin{tabular}{c|cc|c|c|c|c|c|}
\multicolumn{3}{c}{}&\multicolumn{5}{c}{Second Model} \\ \cline{2-8}
&&&\multirow{2}{*}{ZITwBT2}&\multicolumn{4}{c|}{ZITwBT2}\\ \cline{5-8} 
&&&&alr&clr&ilr&clr+PPCA \\ \cline{2-8}  
\multirow{5}{*}{\rotatebox[origin=c]{90}{First Model}}&
\multicolumn{2}{c|}{ZITwBT2}& &- &- &- &- \\ \cline{2-8}
&\multirow{4}{*}{\rotatebox[origin=c]{90}
{ZITwBT2}}&alr&-0.30 (\color{red}0.77)&&-&-&- \\ \cline{3-8}
&&clr&-0.70 (\color{red}0.48)&-0.53 (\color{red}0.60)&&- &-\\\cline{3-8}
&&ilr&-0.88 (\color{red}0.38)&-0.73 (\color{red}0.47)&-0.30 (\color{red}0.76)&&-\\\cline{3-8}
&&clr+PPCA&-0.64 (\color{red}0.52)&-0.46 (\color{red}0.65)&0.14 (\color{red}0.89)&0.42 (\color{red}0.68)& \\ \cline{2-8}  
\end{tabular}} 
\begin{tablenotes}
\scriptsize
\item Numbers represent Vuong statistics, with p-values shown in parentheses.
\end{tablenotes}
\end{threeparttable}
\end{table}

According to \cite{frees2014insurance}, the model selected by the ``min-max'' method stands out as a robust model selection approach, particularly in the context of insurance data analysis. This method involves a strategic selection process aimed at identifying the base model that exhibits the least susceptibility to competing models. In essence, it focuses on identifying the model with the smallest maximum $\text{Gini}^b$ indices across all competing models, thereby highlighting its robustness and stability under various conditions. Table \ref{tab:ginib} presents a comprehensive overview of the results obtained through the ``min-max'' method. Specifically, the ZITwBT2 model emerges with the smallest maximum $\text{Gini}^b$ index recorded at 0.127 among all competing models. This finding further underscores the ZITwBT2 model's robustness and resilience, particularly when compared to alternative models in the analysis.

\begin{table}[htbp]
\caption{$\text{Gini}^{b}$ across 4 Models}\label{tab:ginib}
\centering
\scalebox{1.10}{
\begin{tabular}{c|c|c|c|c|c|}
\multicolumn{2}{c}{}&\multicolumn{4}{c}{Competing Model} \\ \cline{2-6}  
&&TwGLM&TwBT&ZITwBT1&ZITwBT2\\ \cline{2-6} 
\multirow{4}{*}{\rotatebox[origin=c]{90}{Base Model}} &TwGLM&-&0.489&0.120&\color{red}0.504\\ \cline{2-6}
&TwBT&-0.043&-&-0.275&\color{red}0.266\\ \cline{2-6}
&ZITwBT1&0.695&0.598&-&\color{red}0.704\\ \cline{2-6}
&ZITwBT2&\color{blue}0.127&0.035&-0.105&-\\ \cline{2-6}  
\end{tabular}} 
\end{table}

Table \ref{tab:ginib2} presents insightful results regarding the performance of the ZITwBT2 model, particularly in terms of the $\text{Gini}^b$ metric. Upon examination, it becomes evident that the ZITwBT2 model, when implemented without compositional data adjustment, yielded the lowest maximum $\text{Gini}^b$ value among the tested scenarios, reaching a value of 0.128. The recorded $\text{Gini}^b$ value of 0.128 suggests that, contrary to expectations, adjusting for compositional features did not lead to a substantial improvement in the ZITwBT2 model's fit to the telematics data. This finding challenges the assumption that incorporating compositional adjustments inherently enhances model accuracy or predictive capabilities in this context. 

\begin{table}[htbp]
\caption{$\text{Gini}^{b}$ in ZITwBT2 Models with and without Compositional Data Adjustment}\label{tab:ginib2}
\centering
\begin{threeparttable}
\scalebox{1.10}{
\begin{tabular}{c|cc|c|c|c|c|c|}
\multicolumn{3}{c}{}&\multicolumn{5}{c}{Competing Model} \\ \cline{2-8}  
&&&\multirow{2}{*}{ZITwBT2}&\multicolumn{4}{c|}{ZITwBT2}\\ \cline{5-8} 
&&&&alr&clr&ilr&clr+PPCA \\ \cline{2-8}  
\multirow{5}{*}{\rotatebox[origin=c]{90}{Base Model}}&
\multicolumn{2}{c|}{ZITwBT2}&-&\color{blue}0.128&0.122&0.124&0.011\\ \cline{2-8}
&\multirow{4}{*}{\rotatebox[origin=c]{90}{ZITwBT2}}&alr&\color{red}0.160&-&0.145&0.119&0.033\\ \cline{3-8}
&&clr&\color{red}0.762&0.760&-&0.755&0.735\\ \cline{3-8}
&&ilr&0.152&\color{red}0.167&0.138&-&0.059\\ \cline{3-8}
&&clr+PPCA&0.335&\color{red}0.349&0.342&0.304&-\\  \cline{2-8}  
\end{tabular}} 
\end{threeparttable}
\end{table}

To summarize, the results indicate that both the ZITwBT1 and ZITwBT2 models demonstrate comparable accuracy levels and outperform the TwGLM and TwBT models across metrics such as Mean Absolute Deviation (MAD), deviance, and Vuong statistics. Examination of the \(\text{Gini}^{a}\) and \(\text{Gini}^{b}\) measures indicates that the ZITwBT2 model, without any logratio adjustments, shows up as the most reliable choice. Specifically, incorporating methods to adjust compositional data do not significantly enhance model performance. This suggests that Gradient Boosted Decision Tree (GBDT) models exhibit robustness against compositional features and thus offer notable advantages over Generalized Linear Models (GLMs) in terms of convenience and effectiveness. Unlike GLMs which require additional processing steps to handle compositional features, GBDT models maintain high performance without the need for such adjustments.

\subsubsection{Interpretation}\label{subsub:interp}

\begin{figure}[htbp]
\centering
\includegraphics[scale=0.7]{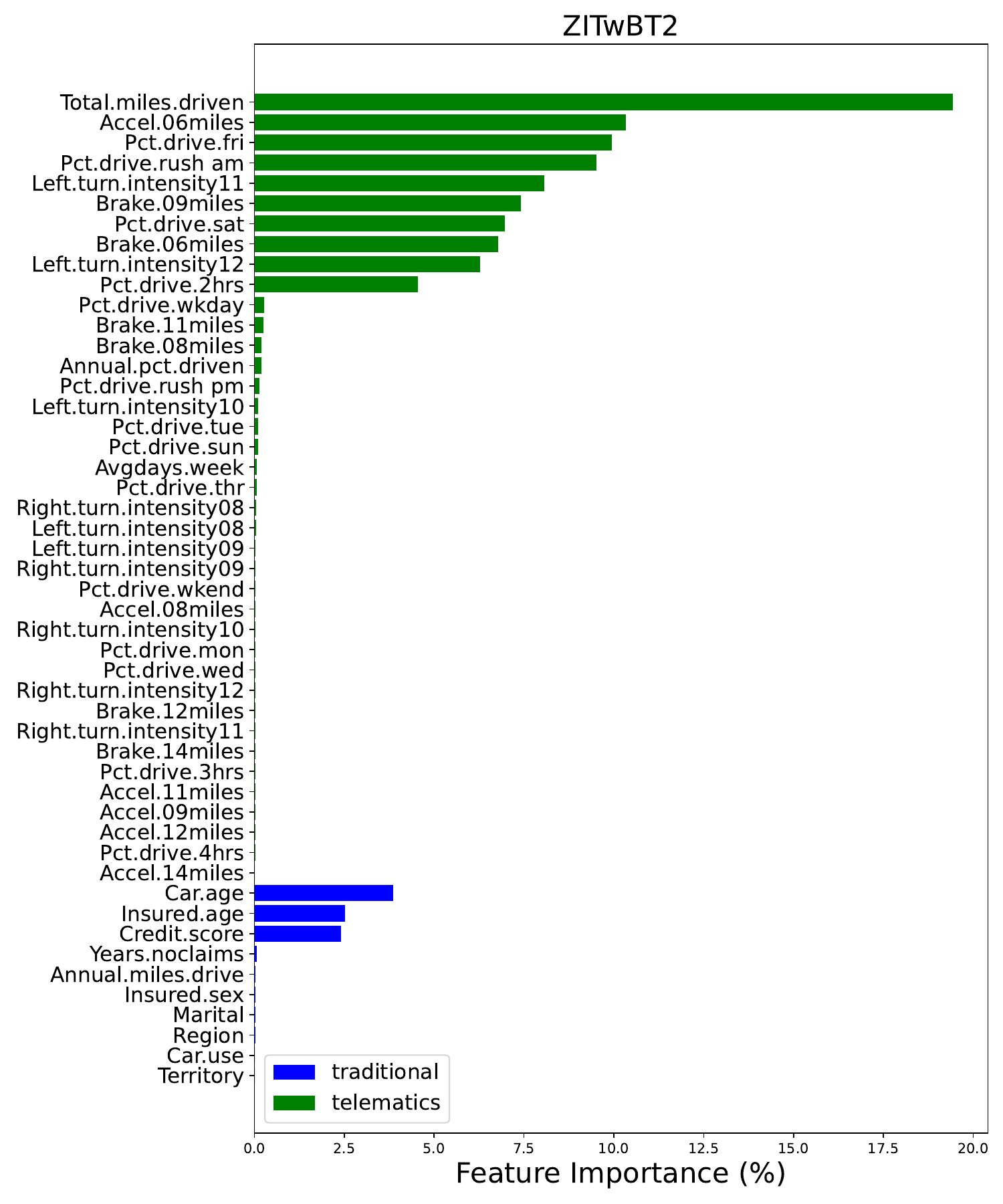}
\caption{Feature Importance in ZITwBT2 \texttt{CatBoost}}	\label{fig3:featureimp}
\end{figure}

Using \texttt{CatBoost} for constructing an aggregate claim amount model offers numerous advantages, especially due to its comprehensive set of model analysis tools that facilitate the interpretation and assessment of diverse risk features' effects and interactions on claim amounts. One of the key methods \texttt{CatBoost} provides is the computation of feature importance values, which plays a crucial role in understanding the model's decision-making process. Feature importance values in \texttt{CatBoost} are calculated based on the impact each feature has on the change in predictions. This impact is measured by examining how distinctively the values of leaves split for a particular feature are distributed. The resulting feature importance values are then normalized, ensuring that the sum of all feature importances equals 100. This normalization allows us for a clearer understanding of the relative importance of each feature in influencing the model's predictions. In Figure \ref{fig3:featureimp}, we present feature importance for the ZITwB2 model, which has been identified as the most reliable model based on prior analyses conducted on synthetic telematics data. In the figure, we ranked feature importance for telematics variables first and later followed by the traditional variables. In particular, the ``Total.miles.driven'' feature holds the highest rank in terms of importance, which indicates its significant impact on predicting aggregate claim amounts. Following closely are features such as ``Accel.06miles,'' ``Pct.drive.fri,'' ``Pct.drive.rush am,'' and ``Left.turn.indensity11,'' all of which are telematics-related features. This underscores the crucial role that telematics features play in determining the overall claim amount, for which it might indicate the importance of leveraging such data in insurance risk assessment and pricing models.

\begin{figure}[htbp]
\centering
\includegraphics[scale=0.65]{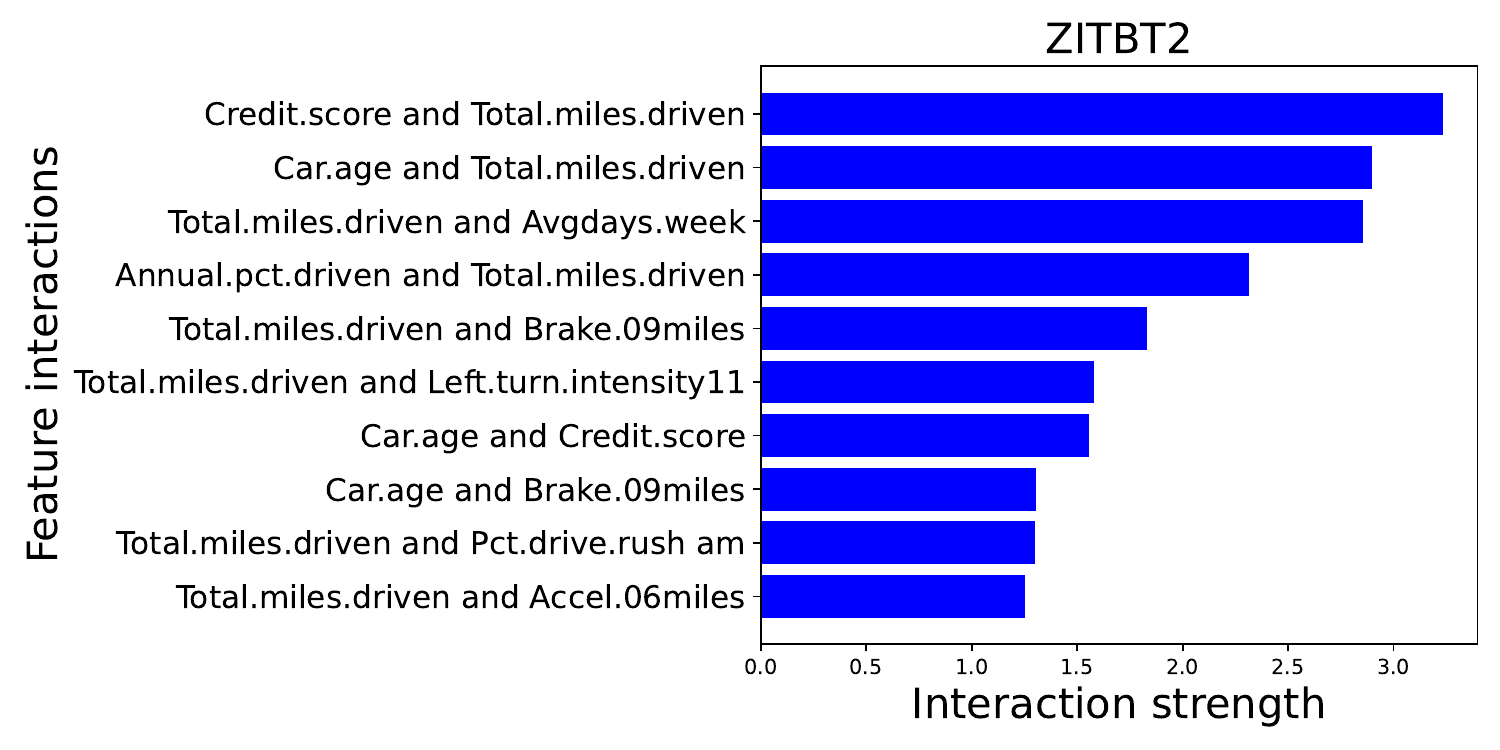}
\caption{Top 10 Features Interaction Strength in ZITwBT2 \texttt{CatBoost} }	\label{fig4:interaction}
\end{figure}

\texttt{CatBoost} provides insights into the strength and impact of relationships between pairs of features. This analysis is particularly valuable for understanding how different features influence each other and contribute to the model's predictive power. When \texttt{CatBoost} calculates interactions between two features, it considers all splits of the feature pair across the trees in the model. This process involves evaluating the magnitude of leaf value changes when the splits of the two features are either aligned in the same direction or differ. Essentially, this approach simulates the effect of fixing one feature while observing how changes in the other feature affect the model's predictions. In practice, understanding feature interactions can significantly enhance our understanding of complex relationships within the data. For example, Figure \ref{fig4:interaction} depicts the top 10 feature interactions identified by the ZITwBT2 model, which was trained on synthetic telematics data. Among these interactions, one of the most prominent ones is observed between the features ``Credit.score'' and ``Total.miles.driven,'' which indicates a strong relationship between an individual's credit score and the total miles they drive. The analysis further reveals other noteworthy interactions where ``Total.miles.driven'' consistently appears paired with features such as ``Car.age,'' ``Avgdays.week,'' and ``Annual.pct.driven.'' These findings provide valuable insights into how driving behavior, vehicle age, and creditworthiness may collectively influence certain outcomes or predictions made by the model. To better understand these interaction effects and their contributions to model predictions, techniques like SHAP (SHapley Additive exPlanations) values can be leveraged. SHAP values help elucidate the impact of individual features and their interactions on model predictions, and this allows for more interpretable and actionable insights from the model.

\texttt{CatBoost}, known for its efficiency in handling categorical variables and robust performance in various machine learning tasks, can be enhanced further by integrating it with the SHAP (SHapley Additive exPlanations) Python library. SHAP values, with roots in game theory principles, provide for a better understanding of feature importance by quantifying each feature's contribution to the model's prediction for a specific observation \cite{lundberg2020local}. When interpreting SHAP values, a positive value for a feature indicates that the feature's presence is pushing the prediction above the average prediction value for that particular observation. Conversely, a negative SHAP value suggests that the feature is pulling the prediction below the average.

\begin{figure}[htbp]
\centering
\includegraphics[scale=0.65]{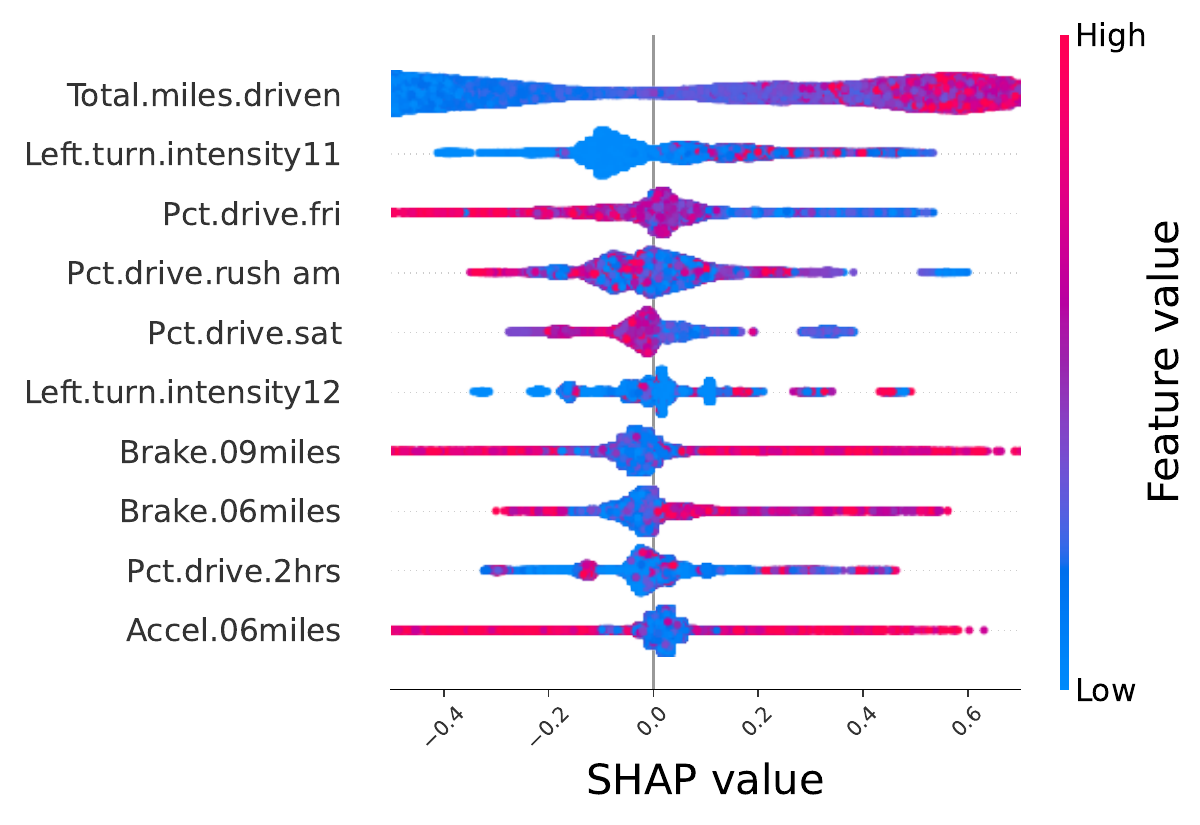}
\caption{SHAP Values of Top Importance Features in ZITwBT2 \texttt{CatBoost}}	\label{fig5:shap}
\end{figure}

In Figure \ref{fig5:shap}, the SHAP values for the top importance features are displayed, which are also depicted in Figure \ref{fig3:featureimp}. These SHAP values provide valuable insights into how different features impact the model's predictions:
\begin{itemize}
\item The feature ``Total.miles.driven'' exhibits a positive correlation with expected claim amounts. This finding implies that individuals who cover longer distances are at a heightened risk of encountering substantial claims, possibly due to increased exposure on the road.
\item Interestingly, increases in the values of ``Pct.drive.fri'' and ``Pct.drive.sat'' are associated with lower expected claim amounts. This observation could be attributed to safer driving practices adopted by individuals during leisure or family-related trips on weekends.
\item Higher values of ``Left.turn.intensity 11'' and ``Left.turn.intensity 12'' correspond to increased expected claim amounts. This suggests that drivers who frequently execute left turns are more likely to incur significant claim amounts. This highlights the elevated risk associated with maneuvers that involve crossing oncoming traffic.
\item For features such as ``Brake.09miles", ``Brake.06miles", and ``Accel.06miles," both extremely high and low values contribute to a greater expected claim amount. This may indicate potential interactions with other features that influence claim predictions.
\end{itemize}

\begin{figure}[htbp]
\centering
\includegraphics[scale=0.43]{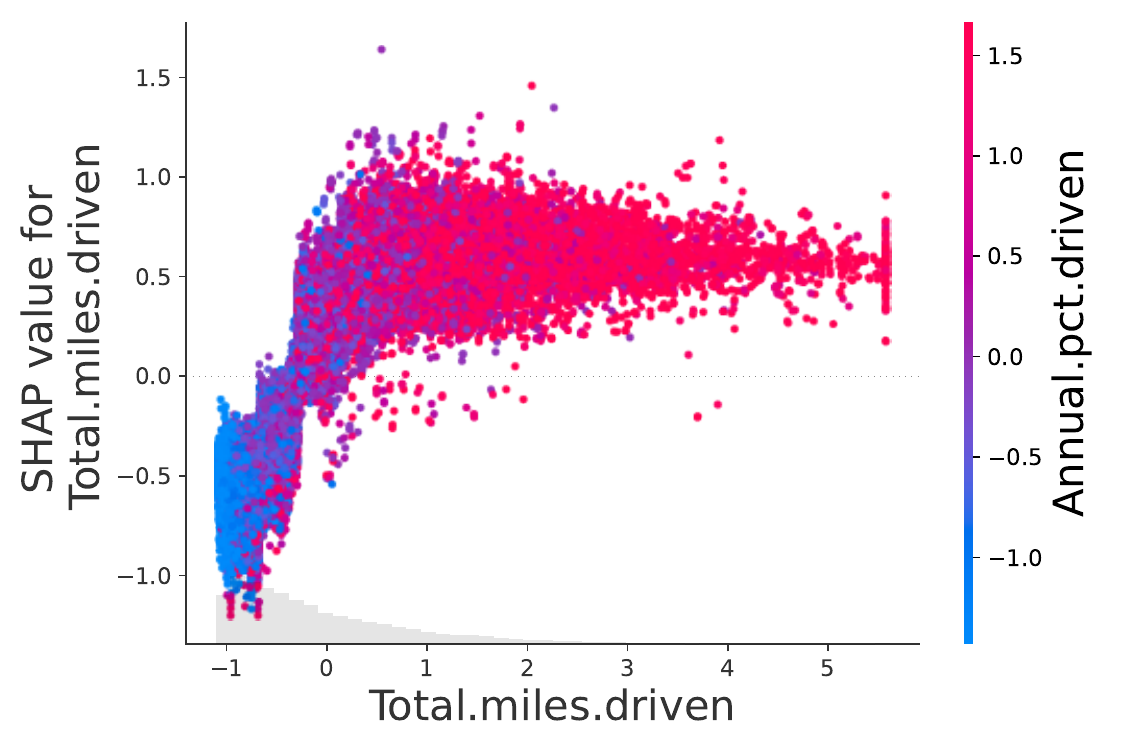}
\includegraphics[scale=0.43]{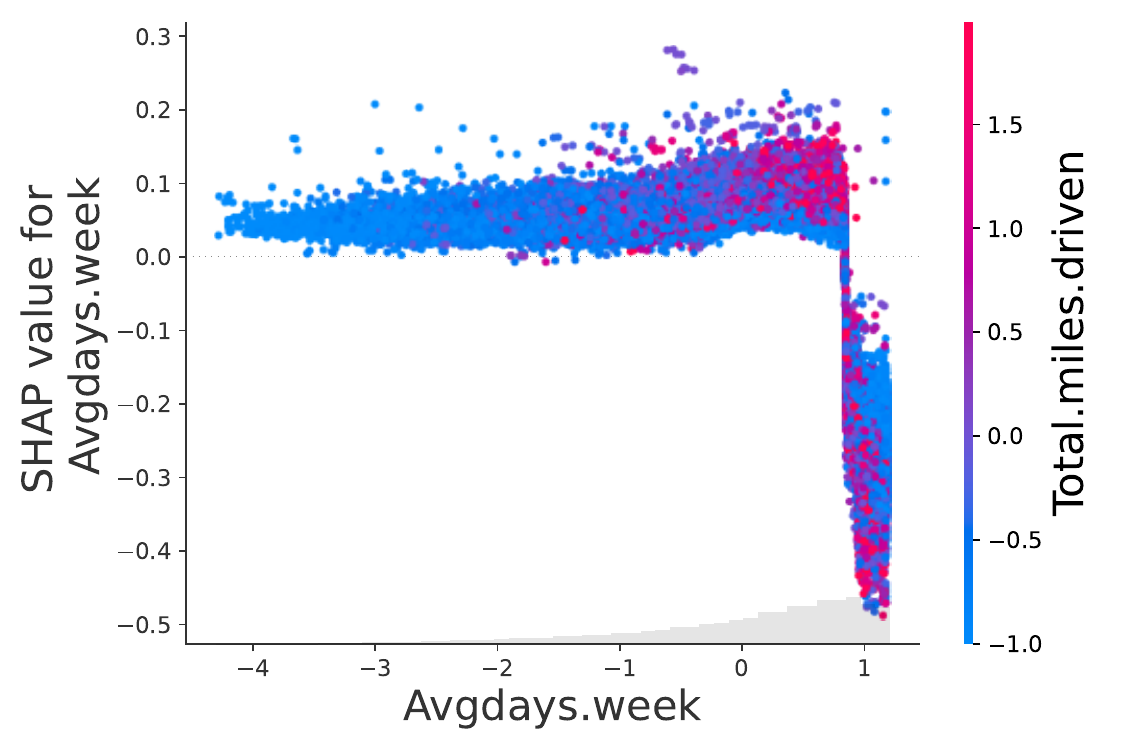}
\includegraphics[scale=0.43]{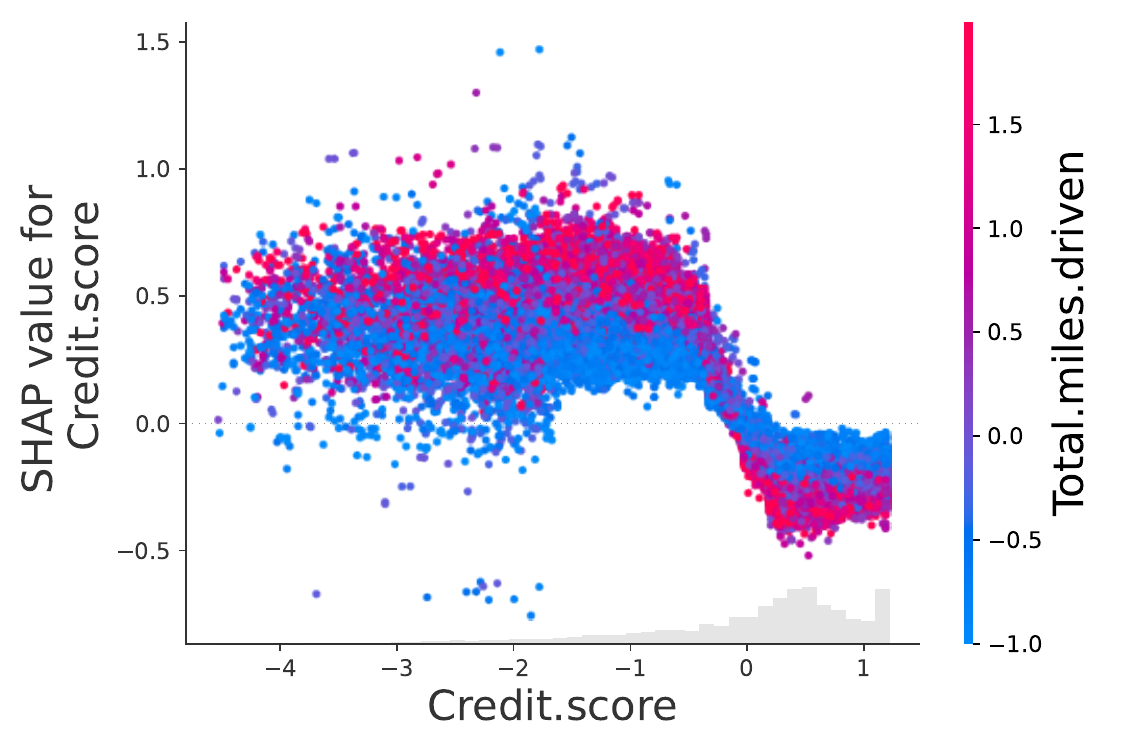}
\includegraphics[scale=0.43]{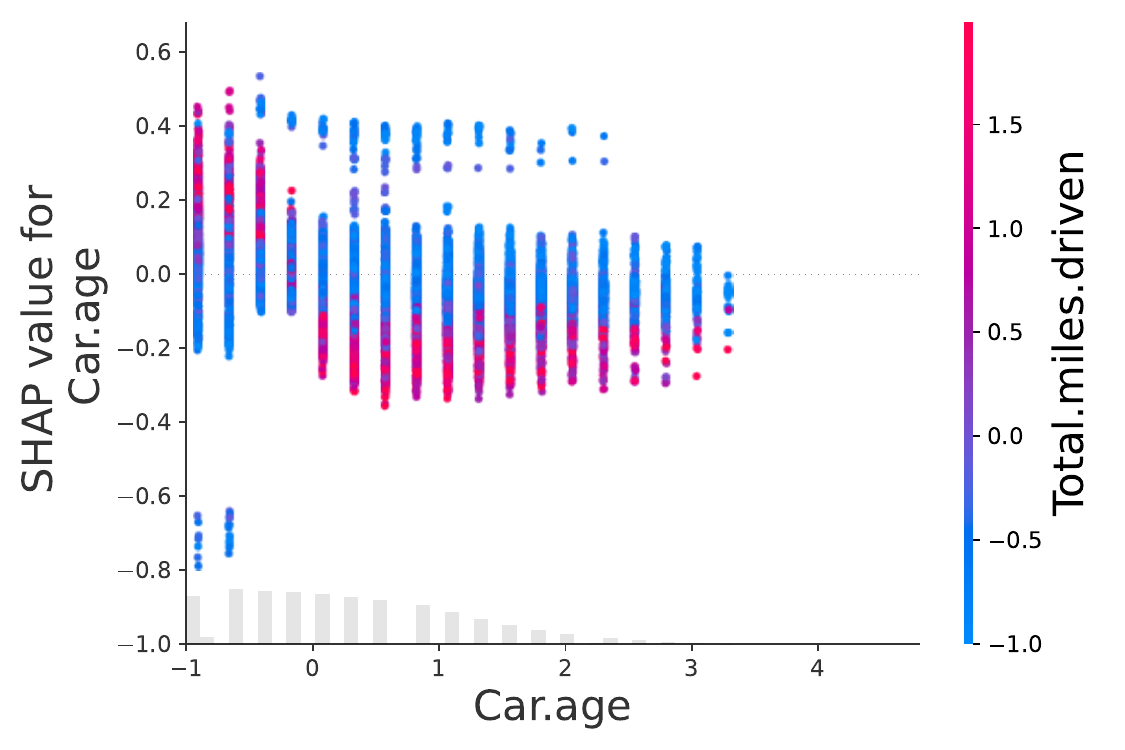}
\includegraphics[scale=0.43]{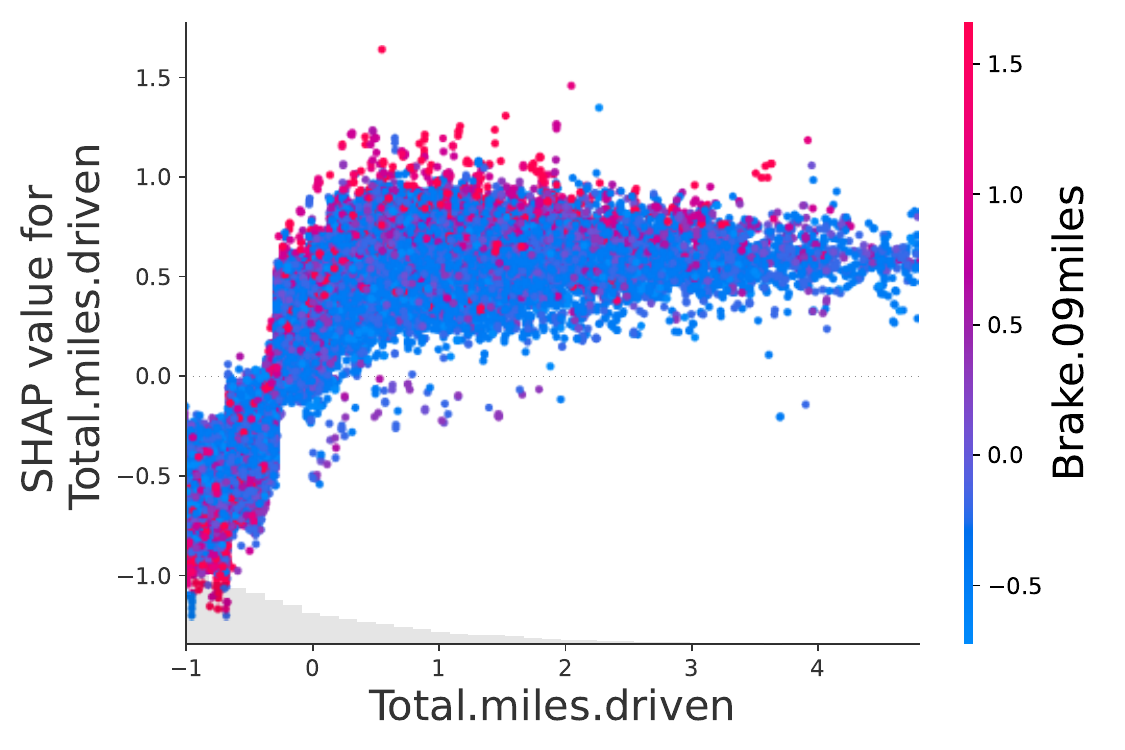}
\includegraphics[scale=0.43]{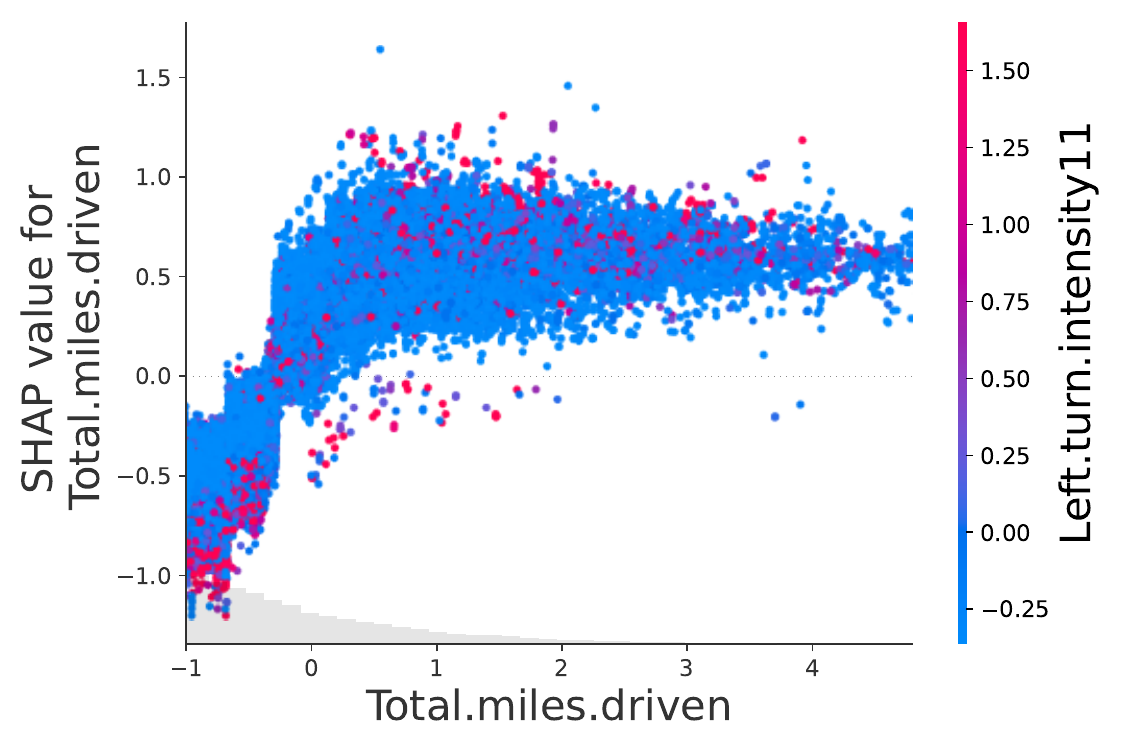}
\caption{Six Scatter Plots Describing Feature Interaction Through SHAP Values}	\label{fig6:shapinteration}
\end{figure}

To gain a deeper understanding of the interaction effects between different features, we utilized scatter plots to visualize the SHAP (SHapley Additive exPlanations) values associated with each feature across all observations. These plots were created with points color-coded by another relevant feature, as illustrated in Figure \ref{fig6:shapinteration}. The strength of interaction between features was assessed and reflected in Figure \ref{fig4:interaction}, resulting in the generation of six distinct plots. In these scatter plots, the x-axis and the right y-axis represent normalized values of the primary feature and the interacting feature, respectively. Meanwhile, the left y-axis corresponds to the SHAP values attributed to the primary feature, providing a comprehensive view of the feature interactions within the dataset.

The plot in the top-left corner of Figure \ref{fig6:shapinteration} illustrates a clear positive correlation between the total annual miles driven and the annualized percentage of time spent on the road. This trend indicates that as both variables increase, there is a corresponding escalation in the expected claim amount. A similar positive correlation is observed between ``Total.miles.driven'' and ``Avgdays.week,'' which provides additional support for the idea that individuals who drive more frequently tend to have higher expected claim amounts. Interestingly, the graph also reveals that if individuals drive nearly every day, regardless of the number of miles driven, the expected claim amount significantly decreases.
 
An analysis of credit scores reveals a clear correlation between lower scores and an elevated risk of substantial insurance claims. Intriguingly, this risk escalates among individuals who log higher mileage. Conversely, higher-than-average credit scores not only mitigate the likelihood of major claims but also indicate safer driving behaviors among those with such scores, which leads to a decrease in claim frequency. Examining the age of vehicles, those below the average age present a notable risk of larger claims, particularly when coupled with extended driving distances. However, as vehicles surpass the average age threshold, the expected claim amount tends to decrease, with a more pronounced decline observed among drivers covering extensive distances. This suggests that longer journeys in older vehicles may potentially contribute to reducing the overall claim amount.

The bottom-left plots in Figure \ref{fig6:shapinteration} reveal an intriguing insight: drivers who frequently cover long distances, brake frequently, and execute numerous left turns tend to have higher insurance claims. On the contrary, individuals who drive extensively but brake infrequently and make fewer left turns are less likely to file large claims. This observation underscores the complex interplay between driving behavior and insurance claim likelihood, particularly with the importance of understanding driving patterns for risk assessment and pricing strategies in the insurance industry.

\begin{figure}[htbp]
\centering
\includegraphics[scale=0.5]{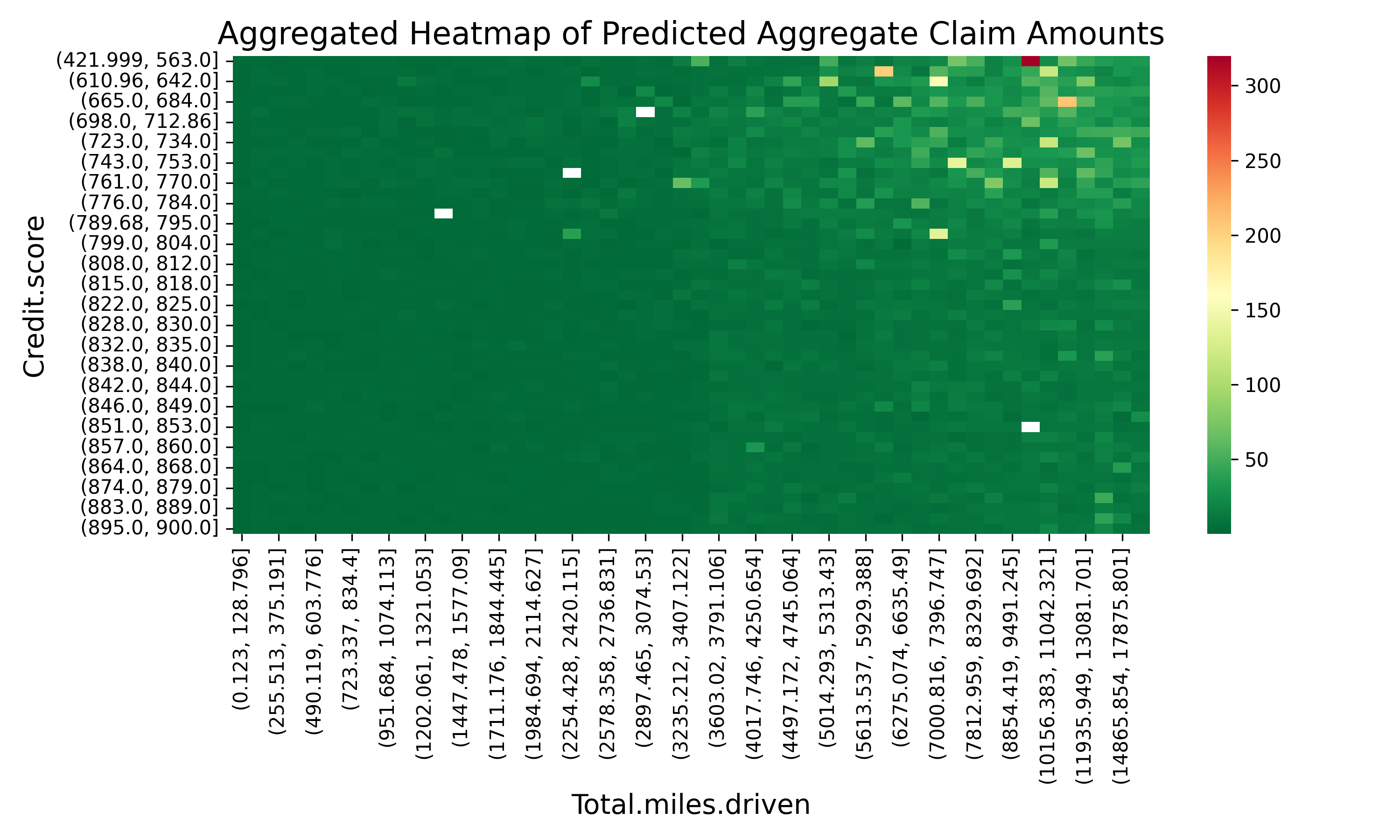}
\includegraphics[scale=0.5]{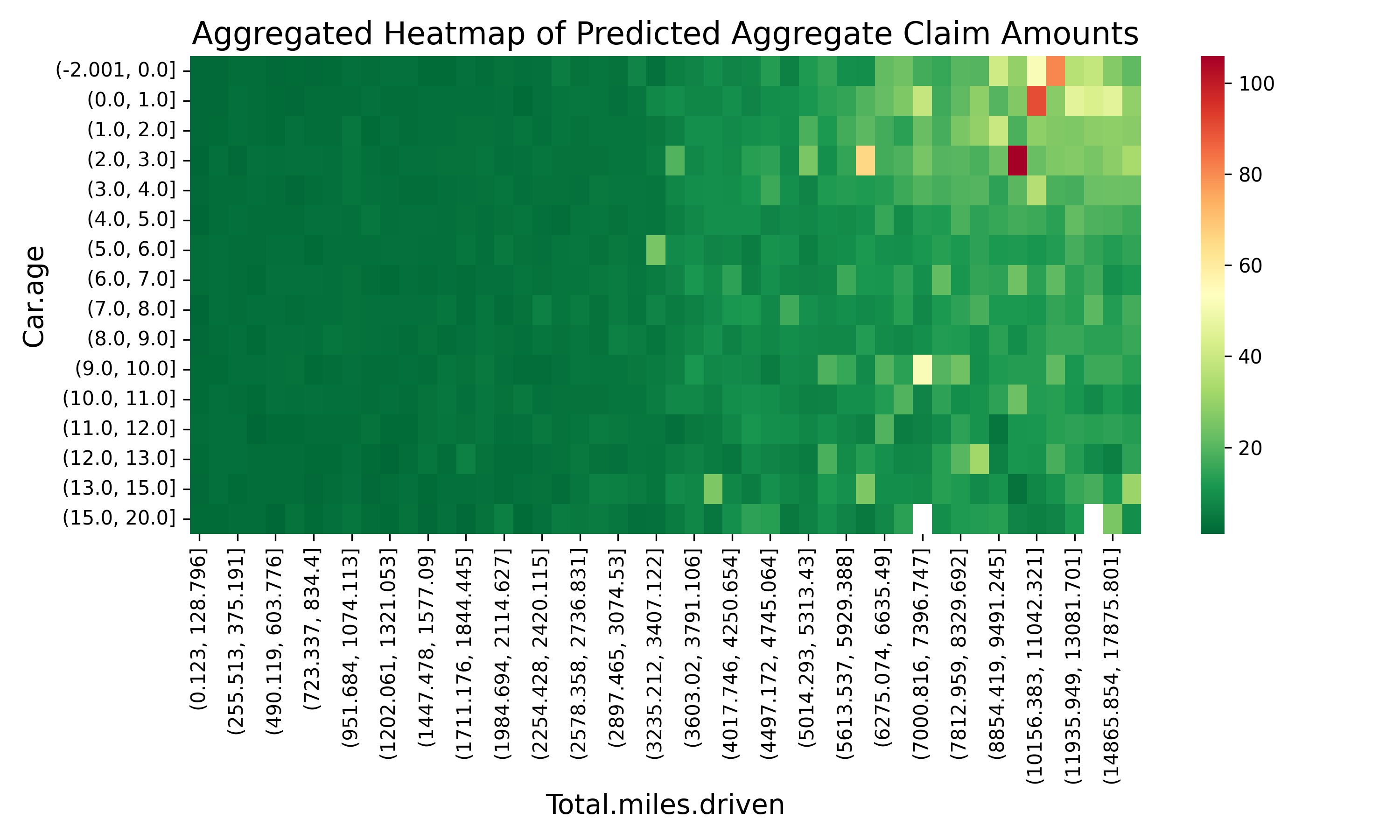}
\includegraphics[scale=0.5]{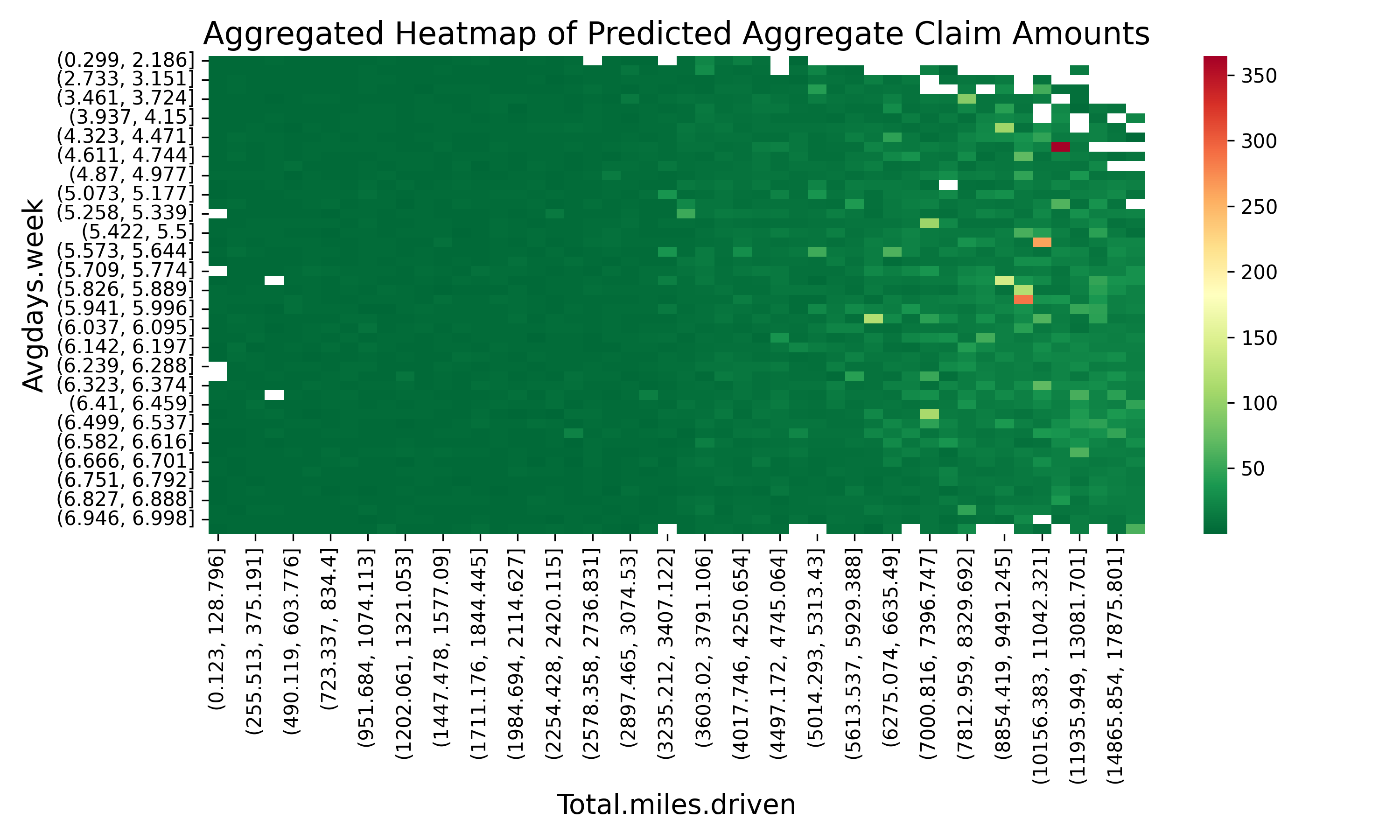}
\caption{Three Heatmaps Describing Feature Interaction with Aggregate Claim Amount}\label{fig7:heatmap}
\end{figure}

To better understand the impact of key feature variables on aggregate claim amounts, we present \ref{fig7:heatmap} that shows a heat map of aggregate claim amounts based on the interaction of ``Total.miles.driven'' and other feature variables. The top graph shows that a low credit score combined with more miles driven can lead to increased claim amounts. In contrast, a high credit score, even with more miles driven, does not result in increased claim amounts. The other two graphs in the figure show similar interactions between miles driven interacting with ``Car.age'' and ``Avgdays.week''. Note that the white cells in the heat map indicate no observations.


\section{Concluding Remarks} \label{sec:conclude}

Broadly speaking, for short-term insurance contracts, it is not uncommon to encounter claim portfolios that exhibit a large mass at zero and a right-skewed distribution for positive claims. This distribution is characterized by a prominent peak at zero, indicating a high likelihood of no claims being made. The adoption of ``zero-inflated'' models, such as the zero-inflated Poisson, negative binomial, and Tweedie regression models, has become more prevalent in practice. While GLM has traditionally been a favored choice in actuarial studies, its linear structure can pose limitations, especially when confronted with nonlinearities and intricate interactions inherent in datasets such as auto telematics data, which captures diverse driving behaviors. Consequently, there is a growing preference for machine learning approaches to better model these complex interactions. Among these approaches, gradient boosting techniques have emerged as powerful tools for constructing predictive models for insurance claims. See \cite{zhou2022-tweedie} and \cite{meng2022actuarial}.

In this paper, we integrated the advantages of applying a zero-inflated Tweedie loss function within a gradient boosting algorithm. To achieve this, we employed several adjustments. Specifically, we reparameterized the zero-inflated Tweedie loss function to enable us to express the inflation probability $q$ as a function of $\mu$. This reparameterization resulted in a singular model, which enabled us to optimally utilize the \texttt{CatBoost} libraries. This approach enhances interpretation and computational efficiency. \texttt{CatBoost} excels at iteratively transforming weak learners into strong learners. Among existing boosting libraries, \citet{so2024enhanced} recommended the use of \texttt{CatBoost} for its efficient implementation in insurance loss estimation.

Importantly, our research demonstrated that boosted models exhibit robustness against the influence of compositional features \citep{aitchison1994principles}, in contrast to GLM models. This robustness implies that boosted models do not require additional adjustments for compositional data, underscoring their superiority over GLMs. Additionally, our study identifies significant risk factors impacting auto insurance claims and meticulously evaluates their individual and combined effects on claim amounts. Leveraging the \texttt{CatBoost} framework in model training has provided dual advantages: efficient handling of categorical features and a comprehensive toolkit for exploring intricate interaction effects embedded within the zero-inflated Tweedie Boosted model. Ultimately, our research illuminates the complex interplay of various features in determining aggregate claim amounts for short-term insurance portfolios.

\clearpage
\subsection*{Appendix A. Detailed steps of ZITwBT1 and ZITwBT2 algorithms} \label{appa-alg}

\begin{algorithm}[H]
	\KwIn{Training dataset $D= \{\bm{x}_i , y_i \}_1^n$, total iterations $T$, learning rate $\alpha$, regularization parameter $\lambda$, power parameter $p$, dispersion parameter $\phi$}
	\KwOut{Final model: $\hat{\mu}_i=w_i \exp(W_T^{mean} (\bm{x}_i))$, $\hat{q}_i=\text{logit}^{-1}(W^{prob}_T(\bm{x}_i))$}
	Initialize trees:  $w^{mean}_0 = 0$, $w^{prob}_0 = 0$, $\phi=1$ \;
	\For{$t=1, \ldots, T$}{
		Compute first(\ref{eq:22}) and second(\ref{eq:23}) derivatives  \hspace{5 in}
		 $g^{mean}_{it}=\partial_{W^{mean}_{t-1}} L(y_i,W^{mean}_{t-1}(\bm{x}_i),W^{prob}_{t-1}(\bm{x}_i))$ \hspace{5 in}
		 $h^{mean}_{it}=\partial^2_{W^{mean}_{t-1}} L(y_i,W^{mean}_{t-1}(\bm{x}_i),W^{prob}_{t-1}(\bm{x}_i))$ for each $i=1,\ldots, n$ \;
		Fit the tree $W^{mean}_T(\bm{x})$ minimizing regularized loss function(\ref{eq:6})\;
		Update $\ W^{mean}_T(\bm{x})=W^{mean}_{t-1}(\bm{x})+\alpha W^{mean}_T(\bm{x})$\;
		
		Compute first(\ref{eq:24}) and second(\ref{eq:25}) derivatives  \hspace{5 in}
		$g^{prob}_{it}=\partial_{W^{prob}_{t-1}} L(y_i,W^{mean}_{t}(\bm{x}_i),W^{prob}_{t-1}(\bm{x}_i))$  \hspace{5 in}
		$h^{prob}_{it}=\partial^2_{W^{prob}_{t-1}} L(y_i,W^{mean}_{t}(\bm{x}_i),W^{prob}_{t-1}(\bm{x}_i))$ for each $i=1,\ldots, n$ \;
		Fit the tree $w^{prob}_t$ minimizing regularized loss function(\ref{eq:6})\;
		Update $\ W^{prob}_t=W^{prob}_{t-1}+\alpha w^{prob}_t$\;
		Update $\phi$ minimizing mean deviation(\ref{eq:50}) \;
	}
	Return final prediction scores:  \hspace{5 in}
	$\ W^{mean}_T(\bm{x}_i) = \sum_{t=1}^{T} \alpha  w^{mean}_t (\bm{x}_i)$, $\ W^{prob}_T(\bm{x}_i) = \sum_{t=1}^{T} \alpha  w^{prob}_t (\bm{x}_i) $\;
	\caption{Zero-Inflated Tweedie Boosted Tree Scenario 1 (ZITwBT1)}\label{alg:case2}
\end{algorithm}

\bigskip
\bigskip

\begin{algorithm}[H]
	\KwIn{Training dataset $D= \{\bm{x}_i , y_i \}_1^n$, total iterations $T$, learning rate $\alpha$, regularization parameter $\lambda$, inflation parameter $\gamma$, power parameter $p$, dispersion parameter $\phi$}
	\KwOut{Final model: $\hat{\mu_i}=w_i \exp(W_T (\bm{x}_i))$, $\hat{q_i}=\displaystyle\frac{1}{1+\hat{\mu_i}^{\gamma}}$}
	Initialize tree:  $w_0 = 0$, $\phi=1$ \;
	\For{$t=1, \ldots, T$}{
		Compute the first(\ref{eq:29}) and second(\ref{eq:30}) derivatives  \hspace{5 in}
		$g_{it}=\partial_{W_{t-1}} L(y_i,W_{t-1}(\bm{x}_i))$  \hspace{5 in}
		$h_{it}=\partial^2_{W_{t-1}} L(y_i,W_{t-1}(\bm{x}_i))$ for each $i=1,\ldots, n$ \;
		Fit the tree $w_t$ minimizing regularized loss function(\ref{eq:6})\;
		Update $\ W_t=W_{t-1}+\alpha w_t$\;
            Update $\phi$ minimizing mean deviation(\ref{eq:50}) \;
            Update $\gamma$ minimizing mean deviation(\ref{eq:50}) \;
	}
	Return final prediction score: $\ W_T(\bm{x}_i) = \sum_{t=1}^{T} \alpha  w_t (\bm{x}_i) $ \;
	\caption{Zero-Inflated Tweedie Boosted Tree Scenario 2 (ZITwBT2)}\label{alg:case1}
\end{algorithm}

\clearpage
\subsection*{Appendix B. Descriptive details of datasets} \label{appb-variables}

\begin{table}[H]
	\centering
	\caption{Variable Names and Descriptions for the Synthetic Telematics Dataset} \label{tab:VD2}
	\begin{threeparttable}
	\resizebox{!}{5.9cm}{
		\begin{tabular}{lll}
			\\
			\toprule
			Type & Variable  & Description \\
			\midrule
			Traditional & Duration & Total exposure in yearly units \\ 
			& Insured.age  & Age of insured driver \\
			& Insured.sex $^\dagger $   & Sex of insured driver: Male, Female \\
			& Car.age  & Age of vehicle (in years) \\
			& Marital $^\dagger $   & Marital status: Single, Married \\
			& Car.use $^\dagger $   & Use of vehicle: Private, Commute, Farmer, Commercial  \\
			& Credit.score  & Credit score of insured driver \\
			& Region $^\dagger $   & Type of region where driver lives: Rural, Urban \\
			& Annual.miles.drive  & Annual miles expected to be driven declared by driver \\
			& Years.noclaims  & Number of years without any claims\\
			& Territory $^\dagger $   & Territorial location of vehicle: 55 labels in \{11, 12, 13, . . ., 91\}\\
			\midrule
			Telematics & Annual.pct.driven  & Annualized percentage of time on the road \\
			& Total.miles.driven  &Total distance driven in miles \\
			&Pct.drive.xxx	&Percent of driving day xxx of the week: mon/tue/…/sun\\
			&Pct.drive.x hrs&Percent vehicle driven within x hrs: 2hrs/3hrs/4hrs\\
			&Pct.drive.xxx	&Percent vehicle driven during xxx: wkday/wkend\\
			&Pct.drive.rush xx	&Percent of driving during xx rush hours: am/pm\\
			&Avgdays.week	&Mean number of days used per week\\
			&Accel.xxmiles	&Number of sudden acceleration 6/8/9/…/14 mph/s per 1000miles\\
			&Brake.xxmiles	&Number of sudden brakes 6/8/9/…/14 mph/s per 1000miles\\
			&Left.turn.intensityxx	&Number of left turn per 1000miles with intensity xx: 08/09/10/11/12\\
			&Right.turn.intensityxx	&Number of right turn per 1000miles with intensity xx: 08/09/10/11/12\\
			\midrule
			Response & NB\_Claim  & Number of claims on the given policy \\
			& AMT\_Claim  & Amount of claims on the given policy \\
			\bottomrule
	\end{tabular}}
 \begin{tablenotes}
	\small
	\item [$\dagger$]  Indicates categorical variable.
\end{tablenotes}
\end{threeparttable}
\end{table}

\clearpage

\bibliographystyle{apalike}
\bibliography{ZIT.bib}

\end{document}